\documentclass[10pt,journal,compsoc]{IEEEtran}
\usepackage{xcolor}
\newtheorem{theorem}{Theorem}[section]

\newtheorem{definition}[theorem]{Definition}
\newtheorem{assumption}[theorem]{Assumption}

\usepackage{hyperref}

\usepackage{pifont}
\usepackage{wrapfig}
\usepackage{diagbox}
\usepackage[T1]{fontenc}

\usepackage{bm}
\usepackage{adjustbox}
\usepackage{booktabs}

\usepackage{amsmath,amsfonts,bm}

\def\eqref#1{equation~\ref{#1}}
\def\1{\bm{1}}

\def\vtheta{{\bm{\theta}}}
\def\vTheta{{\bm{\Theta}}}

\def\vu{{\bm{u}}}

\def\vx{{\bm{x}}}

\def\vz{{\bm{z}}}

\DeclareMathAlphabet{\mathsfit}{\encodingdefault}{\sfdefault}{m}{sl}
\SetMathAlphabet{\mathsfit}{bold}{\encodingdefault}{\sfdefault}{bx}{n}

\def\gA{{\mathcal{A}}}

\def\gD{{\mathcal{D}}}

\def\gF{{\mathcal{F}}}
\def\gG{{\mathcal{G}}}

\def\gL{{\mathcal{L}}}
\def\gM{{\mathcal{M}}}

\def\gO{{\mathcal{O}}}
\def\gP{{\mathcal{P}}}

\def\gS{{\mathcal{S}}}
\def\gT{{\mathcal{T}}}
\def\gU{{\mathcal{U}}}

\def\gX{{\mathcal{X}}}
\def\gY{{\mathcal{Y}}}

\def\sA{{\mathbb{A}}}

\def\sE{{\mathbb{E}}}
\def\sR{{\mathbb{R}}}

\DeclareMathOperator*{\argmin}{arg\,min}

\ifCLASSOPTIONcompsoc
  \usepackage[nocompress]{cite}
\else
  \usepackage{cite}
\fi

\usepackage{enumitem}
\setlist[itemize]{noitemsep, topsep=0pt}

\ifCLASSINFOpdf
\else
\fi

\begin{document}

\title{A Comprehensive Survey of Forgetting in \\ Deep Learning Beyond Continual Learning}

\author{Zhenyi Wang,
        Enneng Yang,
        Li Shen,  
        Heng Huang
\IEEEcompsocitemizethanks{\IEEEcompsocthanksitem Zhenyi Wang and Heng Huang are with the Department
of Computer Science, University of Maryland, College Park,
MD 20742, USA.\protect\\
% note need leading \protect in front of \\ to get a newline within \thanks as
% \\ is fragile and will error, could use \hfil\break instead.
E-mail: wangzhenyineu@gmail.com; henghuanghh@gmail.com
\IEEEcompsocthanksitem Enneng Yang is with Northeastern University, China.  E-mail: ennengyang@stumail.neu.edu.cn
\IEEEcompsocthanksitem Li Shen is with Sun Yat-sen University, China.  E-mail: mathshenli@gmail.com.
}% <-this % stops an unwanted space

% \thanks{Manuscript received July 15, 2023; revised July 15, 2023.}}
\thanks{Manuscript received July 15, 2023; revised June 15, 2024.}}

% The paper headers
\markboth{Journal of \LaTeX\ Class Files,~Vol.~14, No.~8, July~2023}%
{Shell \MakeLowercase{\textit{et al.}}: Bare Demo of IEEEtran.cls for Computer Society Journals}

% Make the directory
% \tableofcontents 

\IEEEtitleabstractindextext{%
\begin{abstract}
Forgetting refers to the loss or deterioration of previously acquired knowledge. While existing surveys on forgetting have primarily focused on continual learning, forgetting is a prevalent phenomenon observed in various other research domains within deep learning. Forgetting manifests in research fields such as generative models due to generator shifts, and federated learning due to heterogeneous data distributions across clients. Addressing forgetting encompasses several challenges, including balancing the retention of old task knowledge with fast learning of new task, managing task interference with conflicting goals, and preventing privacy leakage, etc.
Moreover, most existing surveys on continual learning implicitly assume that forgetting is always harmful. In contrast, our survey argues that forgetting is a double-edged sword and can be beneficial and desirable in certain cases, such as privacy-preserving scenarios. By exploring forgetting in a broader context, we present a more nuanced understanding of this phenomenon and highlight its potential advantages.
Through this comprehensive survey, we aspire to uncover potential solutions by drawing upon ideas and approaches from various fields that have dealt with forgetting. By examining forgetting beyond its conventional boundaries, we hope to encourage the development of novel strategies for mitigating, harnessing, or even embracing forgetting in real applications. 
A comprehensive list of papers about forgetting in various research fields is available at \url{https://github.com/EnnengYang/Awesome-Forgetting-in-Deep-Learning}.

\end{abstract}

}

\maketitle

\section{Introduction}

Forgetting \cite{seq89}  refers to the phenomenon where previously acquired information or knowledge in a machine learning system degrades over time. In the early days of neural networks, the focus was primarily on training models on static datasets. Forgetting was not a significant concern  since the models were trained and evaluated on fixed datasets.
The concept of catastrophic forgetting was first formally introduced by McCloskey and Cohen \cite{seq89}. They demonstrated that neural networks when trained sequentially on different tasks, tend to forget previously learned tasks when new tasks are learned.
Later, addressing the issue of forgetting was formalized as continual learning (CL). Nowadays, forgetting has garnered significant attention not only within the CL domain but also in the broader machine learning community, which has evolved into a fundamental problem in the field of machine learning.

Existing surveys on forgetting have primarily focused on CL \cite{CLsurvey2, vandeven2019scenarios, CLsurvey, MAI202228,  masana2022class, zhou2023deep, composeCL, surveyCL}. However, these surveys tend to concentrate solely on the harmful effects of forgetting and lack a comprehensive discussion on the topic.
In contrast, we highlight the dual nature of forgetting as a double-edged sword, emphasizing both its benefits and harms. Additionally, our survey extends beyond the scope of CL and covers the forgetting issue in various other domains, including foundation models, domain adaptation, meta-learning, test-time adaptation, generative models, reinforcement learning and federated learning. By doing so, we offer a comprehensive examination of forgetting that encompasses a broader range of contexts and applications.

\begin{table*}[t]
\caption{Harmful Forgetting: Comparisons among different problem settings. }
\vspace{-10pt}
\begin{adjustbox}{scale=0.84,tabular= lccccccc,center}
  \begin{tabular}{cccccccccccc}
  \toprule  
\textbf{Problem Setting} &\textbf{Goal} & \textbf{Source of Forgetting}  &  \\ \midrule
Continual Learning & learn non-stationary data distribution without forgetting previous knowledge  & data-distribution shift during training &   \\
Foundation Model &  unsupervised learning on large-scale unlabeled data & data-distribution shift in pre-training, fine-tuning \\
Domain Adaptation& adapt to target domain while maintaining performance on source domain & target domain sequentially shift over time & \\
Test-time Adaptation &mitigate the distribution gap between training and testing & adaptation to the test data distribution during testing\\
Meta Learning  & learn adaptable knowledge to new tasks &incrementally meta-learn new classes / task-distribution shift &\\
Generative Model & learn a generator to approximate real data distribution  & generator shift / data-distribution shift & \\
Reinforcement Learning & maximize accumulate rewards  & state, action, reward and state transition dynamics shift  & \\
Federated Learning & decentralized training without sharing data  &  model average; non-i.i.d data; data-distribution shift & \\
    \bottomrule
  \end{tabular}
\label{tab:harmfulforget}
\end{adjustbox}
\vspace{-10pt}
\end{table*}

\begin{table*}[t]
\caption{Beneficial Forgetting: Comparisons among different problem settings.}
\vspace{-10pt}
\begin{adjustbox}{scale=0.85,tabular= lccccccc,center}
  \begin{tabular}{cccccccccccc}
  \toprule  
\textbf{Problem Setting} & \textbf{Goal}     
\\ 
\midrule
Mitigate Overfitting   & mitigate memorization of training data through selective forgetting\\
Debias and Forget Irrelevant Information & forget biased information to achieve better performance or remove  irrelevant information to learn new tasks \\
Machine Unlearning & forget some specified training data to protect user privacy \\
    \bottomrule
  \end{tabular}
\label{tab:beneficialforget}
\end{adjustbox}
\vspace{-15pt}
\end{table*}
In this survey, we classify forgetting in machine learning into: harmful forgetting and beneficial forgetting, based on the specific application scenarios. Harmful forgetting occurs when we desire the machine learning model to retain previously learned knowledge while adapting to new tasks, domains, or environments. In such scenarios, it is crucial to prevent knowledge forgetting. Conversely, there are many cases where beneficial forgetting becomes necessary. For example: (1) Overfitting to the training data hinders generalization.  (2) Irrelevant and noisy information impedes the model's ability to effectively learn new knowledge. (3) Pre-trained model contains private information that could potentially lead to privacy leakage. In these situations, forgetting becomes desirable as it serves several important purposes. Firstly, forgetting can mitigate overfitting, as it allows the model to forget irrelevant details and focus on the most pertinent patterns in the training data.  Additionally, by discarding unnecessary information, forgetting facilitates the learning of new knowledge, as the model can make better use of its capacity to acquire and adapt to novel information. Lastly, forgetting helps protect privacy by discarding sensitive user information, ensuring that such data is not retained in the model's memory.

\subsection{Harmful Forgetting}

Harmful forgetting has been observed not only in CL but also in various other research areas, including foundation model, domain adaptation, meta-learning, test-time adaptation, generative model, reinforcement learning and federated learning. While existing surveys have predominantly focused on forgetting in CL, this survey aims to fill the gap by providing an overview of forgetting across various learning scenarios.

Forgetting in these research fields can be attributed to various factors. In continual learning, forgetting occurs due to the shift in data distribution across different tasks. In meta-learning, forgetting is a consequence of the shift in task distribution. In federated learning, forgetting is caused by the heterogeneity of data distribution among different clients, commonly known as client drift. In domain adaptation, forgetting happens because of domain shift. In test-time adaptation, forgetting is a result of adapting to the test data distribution during testing. In generative models, forgetting occurs due to the shift in the generator over time or when learning non-stationary data distribution. In reinforcement learning, forgetting can occur as a result of shifts in state, action, reward, and state transition dynamics over time. These changes in the underlying  environment can lead to the loss or alteration of previously learned knowledge in reinforcement learning. In foundation models, forgetting can be attributed to: fine-tuning forgetting, incremental streaming data pre-training, and the utilization of foundation models for downstream CL tasks. 

To facilitate comparisons of various settings related to forgetting, we present a comprehensive analysis of harmful forgetting in Table \ref{tab:harmfulforget}.

\textbf{Harmful Forgetting Definition}. We denote the performance on the test set $X_t$ when learning task/domain $t$ with parameters $\vtheta_t$ as $\gL(\vtheta_t, X_t)$. We denote the performance on the test set $X_t$ after learning the last task/domain $T$ with parameters $\vtheta_T$ as $\gL(\vtheta_T, X_t)$. Forgetting after learning task/domain $T$ can then be defined as follows:
\begin{definition}[\textbf{Forgetting}] 
\begin{equation}
\setlength\abovedisplayskip{3pt}
\setlength\belowdisplayskip{3pt}
F = \frac{1}{T-1}  \sum_{t=1}^{T}(\gL(\vtheta_T, X_t) - \gL(\vtheta_t, X_t))
\end{equation}
\end{definition}
This definition covers various learning scenarios in different settings. For example, for continual learning, $\gL(\vtheta_t, X_t)$ denotes the test set performance on task $t$. For reinforcement learning, $\gL(\vtheta_t, X_t)$ denotes the cumulative reward/average reward/discounted reward on task $t$. For meta-learning, $\gL(\vtheta_t, X_t)$ denotes the meta test set accuracy on task distribution $t$. For generative model, $\gL(\vtheta_t, X_t)$ denotes the Frechet Inception Distance (FID) or Inception Score (IS).

\subsection{Beneficial Forgetting}

 While the prevailing belief in most existing works is that forgetting is harmful, we have come to recognize that forgetting is a double-edged sword. There are many instances where it is advantageous to forget certain knowledge. For example: (1) selective forgetting could help mitigate overfitting; (2) to enhance model generalization or facilitate learning of new tasks/knowledge, it is imperative to eliminate biased or irrelevant information from previously learned knowledge; and (3) machine unlearning, which prevents data privacy leakage.

First, overfitting has remained a fundamental challenge in machine learning, as it arises when a model excessively memorizes the training data but struggles to generalize effectively to new, unseen test data. To improve generalization, it is crucial for the model to avoid the mere memorization of training data and instead should prioritize learning the true underlying relationship between the input data and corresponding labels.
One important technique to enhance generalization is selective forgetting. By selectively discarding irrelevant or noisy information learned from training data, the model can focus on the most pertinent patterns and features, leading to improved generalization performance on unseen data.

Second, when learning new tasks or knowledge, previously acquired knowledge may not always be helpful for improving learning on new information. When a model contains outdated or unrelated knowledge, it can hinder its ability to effectively learn and generalize from new data. In such situations, it is necessary to discard irrelevant information from the model's memory. By freeing up capacity within the model, it becomes more receptive and adaptive to acquiring new knowledge. 
The process of discarding irrelevant information is crucial for preventing interference between old and new knowledge.

Lastly, model users may request the removal of their training data from both the database and the pre-trained model, exercising their Right to Be Forgotten \cite{ginart2019making}. To address this, researchers have developed machine unlearning, which allows models to intentionally forget unwanted private data. Additionally, some privacy attacks exploit the model's tendency to memorize data to extract private information. Membership inference attacks \cite{shokri2017membership} can identify whether a specific data point was part of the training data for a pre-trained model. Intentional forgetting of private data helps protect privacy and prevent information leakage in such cases.

To facilitate comparisons,  we also provide a comparative analysis in Table \ref{tab:beneficialforget} for beneficial forgetting, encompassing the above mentioned diverse settings for reference.

\subsection{Challenges in Addressing Forgetting}

Addressing forgetting faces numerous challenges that vary across different research fields. These challenges include:

\textit{Data Availability}: 
Data availability is a significant challenge for addressing forgetting in various scenarios. Limited access to previous task data, due to storage constraints or privacy concerns, complicates continual learning, meta-learning, domain adaptation, generative models, and reinforcement learning. Additionally, some scenarios, like federated learning, prohibit using raw data, as only the model parameters are shared with a central server.

\textit{Resource Constraints}: Resource-limited environments, such as those with constraints on memory and computation, present challenges in effectively addressing forgetting. In online continual learning and meta-learning, where data or tasks are typically processed only once, these challenges are particularly pronounced. 
Furthermore, online learning often operates in resource-constrained environments with limited memory or computation capabilities. These constraints pose additional hurdles for addressing forgetting. 

\textit{Adaption to New Environments/Distribution}: In continual learning, foundation models, reinforcement learning, domain adaptation, test-time adaptation, meta-learning, and generative models, the target environment or data distribution can change over time. The learning agent must adapt to new scenarios, which can happen during training or testing. However, the agent often forgets previously acquired knowledge or loses performance on earlier tasks due to the data distribution shift.

\textit{Task Interference/Inconsistency}: 
Conflicting goals among different tasks can cause task interference, making it hard to prevent forgetting in continual learning and federated learning. In continual learning, sequential tasks may conflict, making it difficult for the network to balance performance across multiple tasks and exacerbating forgetting. In federated learning, models trained on different clients can show inconsistencies \cite{shi2023improving} due to heterogeneous data distributions, leading to client interference and further worsening the forgetting problem.

\textit{Privacy-Leakage Prevention}: 
In some cases, retaining old knowledge can raise privacy concerns by unintentionally exposing private information. To address these risks and prevent the disclosure of sensitive data, the focus should be on forgetting or erasing training data traces rather than memorizing them. This challenge is central to machine unlearning, which aims to effectively remove training data traces from machine learning models \cite{cao2015towards}.

\subsection{Survey Scope, Contributions and Organization}

\textbf{Survey Scope}. 
Our main objective is to give a comprehensive overview of forgetting in key research areas where it is significant. By exploring these fields, we aim to highlight the existence and impact of forgetting in these domains.

 Our contributions can be summarized into three fold:

\begin{itemize}
\item We provide a more systematic survey on CL compared to existing surveys. Our survey includes a more systematic categorization of CL problem settings and methods.
\item In addition to CL, our survey extends its scope to encompass forgetting in other research fields. This broader coverage provides a comprehensive understanding of forgetting across various research fields.
\item Our survey, in contrast to existing surveys on CL, reveals that forgetting can be viewed as a double-edged sword. We emphasize that forgetting can also have beneficial implications in privacy-preserving scenarios.
\end{itemize}

\textbf{Organization}.
The structure of this paper is as follows.  
In Sections \ref{sec:continuallearning}-\ref{sec:federatedlearning}, we provide a comprehensive survey on harmful forgetting in continual learning, foundation model, domain adaptation, test-time adaptation, meta-learning, generative model, reinforcement learning, and federated learning. Each section explores the occurrence and impact of forgetting within these specific fields.
In Section \ref{sec:beneficial}, we delve into the concept of beneficial forgetting. This section highlights the positive aspects of forgetting in specific learning scenarios. In Section \ref{sec:discussion}, we present the current research trends and offer insights into the potential future developments.

\section{Forgetting in Continual Learning}
\label{sec:continuallearning}

\begin{table*}[!htbp]
\caption{Content outline in CL. Based on different problem setting categorization criteria, the CL setting can be classified into various scenarios, as presented in the following table:}
\vspace{-10pt}
\begin{adjustbox}{scale=1.0,tabular= lccccccc,center}
  \begin{tabular}{cccccccccccc}
  \toprule  
\textbf{Section} & \textbf{Problem Setting}  & \textbf{Categorization Criterion}      
\\ 
\midrule
Section \ref{sec:taskawarefree}   & Task-aware and Task-free CL & whether explicit task splits/information are available or not during training\\
Section \ref{sec:offlineonline} &   Online CL     & the model processes the data in a single pass or multiple passes\\
Section \ref{sec:labelCL} &   Semi-supervised, Few-shot and Unsupervised CL & the amount of labeled data  used in CL  \\
    \bottomrule
  \end{tabular}
\label{tab:setting}
\end{adjustbox}
\vspace{-15pt}
\end{table*}

The goal of continual learning  (CL) is to learn on a sequence of tasks $\gT_1, \gT_2, \cdots, \gT_N$ without forgetting the knowledge on previous tasks. It can be formulated with the following optimization objective. 
Suppose when learning task $t$, the goal is to minimize the risk on all the seen tasks so far, i.e.,
\begin{equation}
\setlength\abovedisplayskip{3pt}
\setlength\belowdisplayskip{3pt}
    \gL(\vtheta_t) = \sum_{t=1}^{N} \sE_{(\vx, y)\sim \gD_{\gT_t}} \gL_{\vtheta_t}(\vx, y),
\end{equation}
where $\sE$ denotes expectation, $\vtheta_t$ are parameters when learning task $t$, and $\gD_{\gT_t}$ represents the training data of task $t$.

The CL problem can be categorized in several different ways. Firstly, according to whether explicit task splits/information are available or not during training stage, CL can be divided into \textit{task-aware (task-based) and task-free (task-agnostic)} scenarios \cite{aljundi2019taskfree}. Task-aware CL can be further classified into task/domain/class incremental learning \cite{vandeven2019scenarios}, depending on whether the task ID is known during testing stage. Among them, task-incremental learning knows the task ID during testing, while domain-incremental learning and class-incremental learning do not know the task ID during the testing phase. In particular, the label space of domain-incremental learning is the same, while other settings have independent label spaces. Addressing forgetting in task-aware CL is relatively straightforward due to the availability of task information.  With knowledge of the specific tasks involved, CL learner can utilize task-specific cues or labels to guide its learning process and manage forgetting.
However, addressing forgetting in task-free CL is more challenging since there are no explicit task splits or task-specific information available. As a result, the learning system must autonomously identify and adapt to changes or shifts in the data distribution without any task-specific cues or labels. This requires the development of robust and adaptive mechanisms that can detect and respond to changes in the data distribution.

Secondly, depending on whether the model processes the data in a single pass or multiple passes, CL can be categorized as \textit{online and offline CL}. Offline CL has been extensively studied due to its availability of abundant computing and storage resources. However, online CL presents unique challenges. In online CL, the agent has limited access to past data and experiences, which restricts the opportunities to revisit and reinforce previously learned tasks. 
Furthermore, online learning often operates in resource-constrained environments with limited memory or processing capabilities. These resource limitations pose additional hurdles for addressing forgetting in online CL.

Lastly, according to the amount of labeled data used in CL, they could be categorized into \textit{supervised, semi-supervised, few-shot, and unsupervised CL}. 
Supervised CL is generally considered the easiest case since the availability of labeled data provides clear task boundaries and evaluation signals. However, challenges arise in other forms of CL.
For semi-supervised CL: the challenge lies in selecting useful knowledge from unlabeled data to mitigate forgetting. Not all unlabeled data may be beneficial for addressing forgetting, making the selection process challenging.
In few-shot CL: the scarcity of labeled data requires the learning agent to effectively utilize the available information to minimize forgetting and adapt to new tasks.
In unsupervised CL: unsupervised CL is the most challenging due to the absence of explicit task boundaries. Defining when a new task begins and differentiating it from previous tasks becomes difficult. 
Furthermore, the absence of labeled data in unsupervised CL results in a scarcity of feedback and evaluation signals for measuring forgetting. 

It is important to note that the terms CL and incremental learning (IL) are often used interchangeably when addressing learning from non-stationary data distributions, as described in \cite{vandeven2019scenarios}. The key objective of CL and IL is to enable models to learn continuously by updating in stages as new data becomes available, ensuring that they can acquire new knowledge while retaining previously learned information. Traditional online learning (OL), by contrast, represents a special case of IL, where the model processes data streams in real time. In OL, the model is updated immediately upon receiving new data, typically handling one sample (or a small batch) at a time from a stationary data distribution/single task. The primary goal of OL is efficient learning rather than mitigating forgetting. This contrasts with CL and IL, where models are updated across multiple epochs and often adapt to changing, non-stationary data distributions.

Below, we present the details of each CL problem setting and its corresponding related works. To make content organization clear, we provide a Table \ref{tab:setting} to summarize the problem setting categorization in the following sections.

\subsection{Task-Aware and Task-Free CL}  
\label{sec:taskawarefree}

\subsubsection{Task-aware CL} 
\label{sec:taskawarecl}

Task-aware CL focuses on addressing scenarios where explicit task definitions, such as task IDs, are available during the CL process. The three most common CL scenarios within task-aware settings are task-incremental learning, domain-incremental learning, and class-incremental learning \cite{vandeven2019scenarios}.
In domain-incremental learning, tasks sequentially arrive with the same label space but different input data distributions. This means that the tasks share a common set of labels or categories, but the distribution of the input data may vary across tasks. 
Task-incremental learning refers to the scenario where tasks arrive sequentially, and each task has its own disjoint label space. During testing, the presence of task IDs allows the model to identify the specific task at hand.
Class-incremental learning does not provide task IDs during testing. Instead, the model needs to incrementally learn new classes without forgetting previously learned classes.

\textit{Problem Setup:} We consider the standard CL problem of learning a sequence of $N$ tasks denoted as $\gD^{tr} = \{\gD_1^{tr}, \gD_2^{tr}, \cdots, \gD_N^{tr}\}$. The training data of $k$-th task $\gD_k^{tr}$ consists of a set of triplets $\{(\vx_i^{k}, y_i^{k}, \gT_k)_{i=1}^{n_k}\}$, where $\vx_i^{k}$ is the $i$-th data example, $y_i^{k}$ is the data label associated with $\vx_i^{k}$, and $\gT_k$ is the task identifier. The goal is to learn a neural network with parameters $\vtheta$, i.e., $f_{\vtheta}$, on $\gD^{tr}$ so that it performs well on the test set of all the learned tasks $\gD^{te} = \{\gD_1^{te}, \gD_2^{te}, \cdots, \gD_N^{te}\}$ without forgetting the knowledge of previous tasks.

\begin{figure*}[t]
\centering  
\begin{adjustbox}{scale=1,tabular= lccccccc,center}
\includegraphics[width=1.\textwidth]{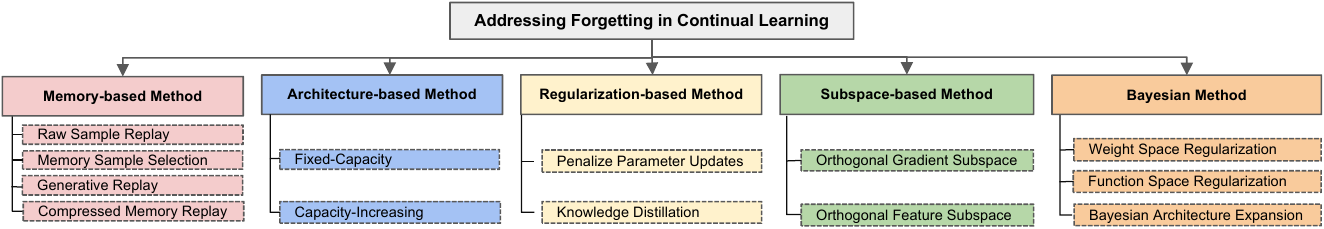}
\end{adjustbox}
\vspace{-15pt}
\caption{Categorization of existing continual learning approach.
}
\vspace{-15pt}
\label{fig:cl_classification}
\end{figure*}

Existing methods on task-aware CL have explored five main branches: memory-based, architecture-based, regularization-based, subspace-based, and Bayesian-based methods. An overview of these branches is provided in Figure \ref{fig:cl_classification}. For a more comprehensive description of the methods within each category, please refer to Appendix \ref{sec:taskawarecl_appendix}. Below, we provide a brief description of each class method.

\paragraph{\textbf{Memory-based Method}} 
Memory-based method keeps a \emph{memory buffer} that stores the data examples from previous tasks and replays those examples during learning new tasks. It can be further categorized into: raw memory replay; memory sample selection; generative replay; and compressed memory replay. Next, we discuss each direction in detail.
(1) \textit{Raw Sample Replay:} These methods randomly save a small amount of raw data from previous tasks and train the model together with the new task data. When the new task updates the model, the old task data is used as a constraint~\cite{lopezpaz2017gradient,AGEM19} or directly mixed with the new data to form a batch~\cite{buzzega2020dark} to update the model, thereby alleviating forgetting.
(2) \textit{Memory Sample Selection:} Randomly selecting samples for replay ignores the amount of information in each sample, which can lead to suboptimal performance~\cite{GCR_CVPR2022,InfluenceCL_CVPR2023}. Therefore, \textit{heuristic selection} selects samples to be stored according to certain rules. For example, select the representative sample closest to the cluster center~\cite{icarl}, the samples with higher diversity~\cite{aljundi2019gradient,bang2021rainbow}, or the difficult sample closer to the decision boundary~\cite{chaudhry2018riemannian,NIPS2019_9357}. 
(3) \textit{Generative Replay:} When privacy concerns restrict the storage of raw memory data, generative replay provides an alternative approach in CL to replay previous task data. The main concept behind generative replay is to train a generative model capable of capturing and remembering the data distribution from previous tasks. The representative works  include GAN-based~\cite{chenshen2018memory,xiang2019incremental}, AutoEncoder-based~\cite{kemkerfearnet}, Diffusion-based~\cite{cil_diffusion_Arxiv2023}, and Model-inversion~\cite{AlwaysBeDreaming_ICCV2021}.
(4) \textit{Compressed Memory Replay:} In scenarios with strict storage constraints on edge devices, memory efficiency becomes a critical consideration. Existing works store feature representations~\cite{van2020brain_naturecommunications2020,BiRT_ICML2023} or low-fidelity images~\cite{wang2022memory,Luo2023ClassIncremental} instead of original images, or learning a set of condensed images~\cite{Mnemonics_liu2020mnemonics,dengremember2022,yang2023an} using dataset distillation~\cite{cazenavette2022dataset}.

\paragraph{\textbf{Architecture-based Method}}  
Architecture-based methods in CL \cite{rusu2016progressive_pnn, fern2017pathnet, yoon2018lifelong} involve updating the network architecture during the learning process to retain previously acquired knowledge. These methods aim to adapt the model's architecture to acquire new tasks while preserving the knowledge from previous tasks.
Based on whether the model parameters expand with the number of tasks, architecture-based methods can be categorized into two types: fixed-capacity and capacity-increasing methods.
(1) \textit{Fixed-Capacity}: In these methods, the amount of CL model's parameters does not increase with the number of tasks, and each task selects a sub-network from the CL model to achieve knowledge transfer and reduce the forgetting caused by sub-network updates. Common subnetwork selection techniques include masking~\cite{HAT2018,bellecdeep,wangsparcl2022}, and pruning~\cite{mallya2018piggyback,mallya2018packnet,hung2019compactingNeurIPS}. 
(2) \textit{Capacity-Increasing}: As the number of tasks increases, fixed-capacity CL models may face limitations in accommodating new tasks. To overcome this challenge, dynamic capacity methods are proposed~\cite{rusu2016progressive_pnn,yoon2018lifelong,L2grow,Der_CVPR2021}. These methods ensure that old tasks are not forgotten and adapt to new tasks by introducing new task-specific parameters for each new task, while freezing parameters related to old tasks.

 \paragraph{\textbf{Regularization-based Method}} 
These methods in CL involve the addition of regularization loss terms to the training objective to prevent forgetting previously learned knowledge \cite{EWC16,zenke2017continual,oswald2019continual}. 
It can be further divided into two subcategories: penalizing important parameter updates and knowledge distillation using a previous model as a teacher.
(1) \textit{Penalize Parameter Updates:} These methods use the Fisher information matrix~\cite{EWC16} or the cumulative update amount of parameters~\cite{MAS2018} as a measure of the importance of old task parameters. On the one hand, when new tasks update important parameters, a large penalty is imposed in order to keep the knowledge of old tasks from being forgotten. On the other hand, imposing a small penalty on unimportant parameter updates helps learn new task's knowledge~\cite{zenke2017continual,UCL_NeurIPS2019,chaudhry2018riemannian}.
(2) \textit{Knowledge-Distillation-Based:} Several methods in CL incorporate a knowledge distillation ~\cite{hinton2015distilling} loss between the network of the previous task (referred to as the teacher) and the network of the current task (referred to as the student) to mitigate forgetting~\cite{LwF,dhar2019learninglwm,bmc_cvpr2023}. It should be mentioned that the ideal scenario would involve using raw data from old tasks to extract the knowledge of the teacher model and refine it into the student model. However, accessing raw data of old tasks is often not feasible due to data privacy concerns. Therefore, existing methods utilize proxy data, such as new task data~\cite{LwF} or large-scale unlabeled data~\cite{lee2019overcoming}, as a substitute for distillation.

\paragraph{\textbf{Subspace-based Method}} Subspace-based methods in CL aim to address the issue of interference between multiple tasks by conducting learning in separate and disjoint subspaces, thus reducing old task forgetting. Subspace-based methods can be categorized into two types based on how the subspaces are constructed: 
% orthogonal gradient subspace and orthogonal feature subspace methods.
(1) \textit{Orthogonal Gradient Subspace}: These methods require that the parameter update direction of the new task is orthogonal to the gradient subspace of the old tasks~\cite{OGDCL,PCAOGD_AISTATS2021,ChaudhryOrthogSubspaceCL}, ensuring minimal interference between tasks.
(2) \textit{Orthogonal Feature Subspace}: These require that the parameter update direction of the new task is orthogonal to the subspace spanned by the input(feature) of the old tasks~\cite{owm,saha2021gradient,deng2021flattening,lin2022trgp,yang2023data}.

\begin{figure}
\vspace{-10pt}
\centering  
\begin{adjustbox}{scale=.45,tabular= lccccccc,center}
    \includegraphics[width=1.\textwidth]{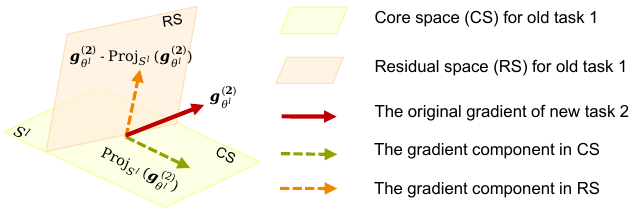}
\end{adjustbox}
    \vspace{-13pt}
\caption{An illustration of the principle of orthogonal projection method.}
    \vspace{-15pt}
\label{fig:gradient_projection}
\end{figure}

We illustrate the working principle of the subspace-based methods (i.e., the orthogonal projection methods) in Fig.~\ref{fig:gradient_projection}. 
Specifically, we define the core subspace (CS) spanned by task 1 in the $l^{th}$ layer as $\small S^l$, constructed from the gradients or features of task 1. The orthogonal space of the core subspace is denoted as the residual subspace (RS). When a new task 2 updates the network in the $l^{th}$ layer with parameters $\small \vtheta^l$, the original gradient direction $\small \mathbf{g}_{\vtheta^{l}}^{(2)}$ is decomposed into CS and RS components. Only the gradient component in the RS, given by $\small \mathbf{g}_{\vtheta^{l}}^{(2)}-\text{Proj}_{S^l}(\mathbf{g}_{\vtheta^{l}}^{(2)})$, is used to update the parameter $\small \vtheta^l$.
Next, we demonstrate why subspace-based approaches (e.g., \cite{saha2021gradient, lin2022trgp, sgp_aaai2023}) can alleviate the forgetting problem. Let the network's weight after training on task 1 be $\small \vtheta_1^l$, and the weight update for task 2 be $\small \Delta \vtheta_2^l$, resulting in the network's weight after training on task 2 as $\small \vtheta_2^l = \vtheta_1^l + \Delta \vtheta_2^l$.
Clearly, the input sample $\small \boldsymbol{x}_{1, i}^l$ from task 1 lies in the subspace $\small S^l$ spanned by task 1. Since the update for task 2 is performed along the subspace orthogonal to $\small S^l$, we have $\Delta \vtheta_2^l \boldsymbol{x}_{1, i}^l=0$. In other words, if task 2 updates are restricted to the direction orthogonal to the subspace spanned by task 1, we ensure that $\small \vtheta_2^l \boldsymbol{x}_{1, i}^l = \vtheta_1^l \boldsymbol{x}_{1, i}^l$, thereby preventing forgetting. The formal statement is as follows:
\begin{equation}
\small
\setlength\abovedisplayskip{3pt}
\setlength\belowdisplayskip{3pt}
    \vtheta_2^l \boldsymbol{x}_{1, i}^l=\left(\vtheta_1^l+\Delta \vtheta_2^l\right) \boldsymbol{x}_{1, i}^l=\vtheta_1^l \boldsymbol{x}_{1, i}^l+\Delta \vtheta_2^l \boldsymbol{x}_{1, i}^l=\vtheta_1^l \boldsymbol{x}_{1, i}^l.
\label{eq:gpm}
\end{equation}

In Appendix~\ref{sec:subspace_appendix}, we discuss in detail the respective advantages and disadvantages of gradient projection in feature space and gradient space, and how to choose them.

\paragraph{\textbf{Bayesian Method}} Bayesian approaches offer effective strategies to mitigate forgetting by incorporating uncertainty estimation and regularization techniques, thereby enhancing the adaptability of the learning process. Bayesian methods can be classified into three categories: (1) constrain the update of weight parameter distributions; (2) constrain the update in function space; and (3) dynamically grow the CL model architecture in an adaptive and Bayesian manner.
Specifically,
(1) \textit{Weight Space Regularization:} These methods model the parameter update uncertainty and enforce the model parameter (weight space) distribution when learning the new task is close to that of all the previously learned tasks, including \cite{nguyen2017variational,IMM_NeurIPS2017,Adel2020Continual, Kurle2020Continual, henning2021posterior, kao2021natural}. 
(2) \textit{Function Space Regularization:} Different from weight space regularization which constrains the weight update, the function space regularization regulates the CL function update in the function space.
They achieve this goal by enforcing the posterior distribution over the function space \cite{pan2020continual}, constraining neural network predictions \cite{titsiasfunctional2020}, modeling the cross-task covariances \cite{kapoor2021variational} or sequential function-space variational inference \cite{rudner2022continual}.
(3) \textit{Bayesian Architecture Expansion:} Bayesian architecture growing methods employ a probabilistic and Bayesian approach to dynamically expand the CL model. This probabilistic framework facilitates the flexible and principled expansion of the model's architecture, allowing it to accommodate increasing complexity and variability in the learning process, including \cite{kumar2021bayesian, mehta2021continual}.

In Appendix \ref{sec:appendix_bayesian}, we discuss the differences and connections between weight space regularity and function space regularity in detail.

\subsubsection{Task-free CL}
\label{sec:taskfreecl}

Task-free CL assumes that the learning system does not have access to any explicit task information. Unlike the task-aware CL setting, where a sequence of tasks is defined, task-free CL aims to perform adaptation without explicit task boundaries.
The system needs to adapt and generalize its knowledge over time, continually updating its model or representation to accommodate new information while retaining previously learned knowledge.
Existing approaches for task-free CL can be categorized into two classes: memory-based methods and network expansion-based methods.

\paragraph{\textit{Memory-based method}} 
Memory-based methods \cite{aljundi2019gradient, NIPS2019_9357, ERRing19, AGEM19} involve storing a small subset of previous data and replaying them alongside new mini-batch data. MIR \cite{NIPS2019_9357} selects and replays samples that are most prone to interference. This selective replay aims to prioritize samples that are most relevant for retaining previously learned knowledge. Building upon MIR, GEN-MIR \cite{NIPS2019_9357} incorporates generative models to synthesize memory examples during replay. GSS~\cite{aljundi2019gradient} focuses on storing diverse examples.  GMED~\cite{jin2021gradientbased}, proposes a method for editing the memory examples to promote forgetting and discourage memorization. While GMED focuses on editing memory examples, Wang et al. \cite{pmlr-v162-wang22v} propose a Distributionally Robust Optimization framework that considers population- and distribution-level evolution to address memory overfitting.

\paragraph{\textit{Expansion-based method}} 
In architecture expansion-based methods, several approaches have been proposed to address the forgetting issue and facilitate continual adaptation.
CN-DPM \cite{lee2020neural} introduces a method that expands the network structure based on the Dirichlet process mixture model. 
This approach allows for the automatic expansion of the network to accommodate new data distributions or concepts while preserving previously learned knowledge. VariGrow \cite{pmlr-v162-ardywibowo22a} proposes a variational architecture growing method based on Bayesian novelty to identify novel information and dynamically expand the network architecture to accommodate new knowledge. ODDL \cite{ye2022taskfree} proposes a dynamical architecture expansion method based on estimating the discrepancy between the probabilistic representation of the memory buffer data and the accumulated knowledge.

\subsection{Online CL} \label{sec:offlineonline}

\subsubsection{Approaches for Online CL} 
 In online CL, the learner is only allowed to process the data for each task once ~\cite{MAI202228}. Existing works addressing forgetting in online CL are mainly based on \textit{rehearsal replay}~\cite{NIPS2019_9357,aljundi2019gradient,onlineshapley,pmlr-v162-guo22g,wumitigating}.
MIR \cite{NIPS2019_9357} suggests replaying the samples that exhibit the maximum increase in loss. OCS~\cite{OCS_ICLR2022} proposes to select samples with high affinity for old tasks. 
DVC~\cite{DVC_CVPR2022} introduces a technique that involves selecting samples whose gradients are most interfered with new incoming samples to store in memory buffer. 
ASER \cite{onlineshapley} introduces an adversarial Shapley value method to score memory data samples based on their contribution to forgetting
La-MAML~\cite{gupta2020maml} utilizes a meta-learning algorithm to tackle online CL by leveraging a small episodic memory. 
POCL~\cite{wumitigating} reformulates replay-based online CL into a hierarchical gradient aggregation framework and enhances past task performance while maintaining current task performance using Pareto optimization.
In addition, some studies have proposed  \textit{regularization-based strategies} to prevent forgetting~\cite{BLD_ECCV2020, PandC_icml2018}.

\subsubsection{Imbalanced Class Issue in Online CL} 
The presence of imbalanced data streams in online CL has drawn significant attention, primarily due to its prevalence in real-world application scenarios~\cite{castro2018end, prabhu2020gdumb, wu2019largebic, ahn2021ss}. Addressing class imbalance can be approached through two main strategies: (1) learning a model that effectively balances the learning of both old and new classes during training, or (2) employing post-processing techniques to calibrate the biases inherent in the model.

\paragraph{\textit{Balance Learning Between New and Old Classes}} 
Balancing in the training phase involves heuristically selecting a balanced memory to tune the model~\cite{castro2018end,prabhu2020gdumb,imbalance2020,kim2020imbalanced}. Chrysakis et al.~\cite{imbalance2020} propose class-balancing reservoir sampling (CBRS) to tackle this issue. PRS \cite{kim2020imbalanced} suggests a partitioning reservoir sampling strategy to address this issue. Kim et al.~\cite{sun2022informationtheoretic} introduces a stochastic information-theoretic reservoir sampler to select memory points from the imbalanced data stream. E2E~\cite{castro2018end} proposes to alleviate the imbalance problem by adopting a balanced fine-tuning strategy at the end of each incremental stage. GDumb~\cite{prabhu2020gdumb} found that the downsampling strategy can well solve the problem of imbalance between old and new classes. 

\paragraph{\textit{Post-Processing Calibration Techniques}} 
Post-processing calibration methods perform bias calibration on the classifier during inference phase~\cite{wu2019largebic,IL2MCV,zhao2020maintaining}. BiC~\cite{wu2019largebic} introduces a two-stage training where they perform the main training in the first stage, followed by a linear transformation to mitigate bias in the second stage. WA~\cite{zhao2020maintaining} reduces the imbalance between old and new classes by aligning the model logits outputs on the old and new classes.
OBC~\cite{OnlineBiasTFCL2023} provides both theoretical and empirical explanations of how replay can introduce a bias towards the most recently observed data stream. They modify the model's output layer, aiming to mitigate the online bias.

\subsection{Semi-supervised, Few-shot and Unsupervised CL}  \label{sec:labelCL}

\subsubsection{Semi-supervised CL}
\label{sec:semi-supervisedcl}

Semi-supervised CL is an extension of traditional CL that allows each task to incorporate unlabeled data as well. 
Existing works mainly include generative replay~\cite{ren2022semi_SIGKDD2022,semi-CL} and distillation~\cite{lee2019overcoming,semi-CL2} to avoid forgetting. Specifically, 
ORDisCo~\cite{semi-CL} maintains a relatively constant-sized network, and it simultaneously trains a classifier and a conditional GAN, and learns the classifier by replaying data sampled from the GAN in an online fashion. SDSL~\cite{ren2022semi_SIGKDD2022} is also based on the generation-replay framework. GD~\cite{lee2019overcoming} and DistillMatch~\cite{semi-CL2} are distillation-based approaches. DistillMatch performs knowledge distillation by assigning pseudo-labels and data augmentation to the unlabeled data. In particular, DietCL~\cite{zhang2024continual} explores semi-supervised CL scenarios with sparse labeled data and limited computational budget.

\subsubsection{Few-shot CL}
\label{sec:fewshotcl}

Few-shot CL refers to the scenario where a model needs to learn new tasks with only a limited number of labeled examples per task while retaining knowledge from previously encountered tasks. The challenge lies in effectively leveraging the limited labeled data and previously learned knowledge to adapt to new tasks while avoiding forgetting.

Compared to traditional CL, few-shot CL faces the challenge of overfitting due to the limited number of examples available per task~\cite{yang2022dynamic_tpami,tao2020few_cvpr2020}. To tackle the forgetting problem in few-shot CL, existing approaches employ various techniques, including metric learning, meta-learning, and parameter regularization. Due to limited pages, we provide details of these methods in Appendix.~\ref{sec:fewshotcl_appendix}. 
Below, we briefly explain each method:
(1) \textit{Metric Learning-Based:} These methods perform classification by class prototypes. To avoid forgetting, the prototype of the new class should be separable from the old class~\cite{zhu2021self_cvpr,hersche2022constrained_cvpr,yangneural_2023_iclr}, and the prototype of the old class should not change drastically during the adjustment process of the new class~\cite{yangneural_2023_iclr,yang2022dynamic_tpami}.
(2) \textit{Meta-Learning-Based:} These methods simulate the inference phase during training so that CL models can quickly adapt to unseen new classes to solve few-shot CL. For example, LIMIT~\cite{zhou2022few_tpami} and MetaFSCIL~\cite{chi2022metafscil_cvpr} split the base task into multiple 'fake'-incremental tasks, so that the model has the learning ability of few-shot CL tasks. By reducing the loss associated with the meta-objective, they minimize forgetting of the old tasks.
(3) \textit{Parameter Regularization-Based:} These methods employ various strategies to address the forgetting problem by penalizing parameter updates that are important for old tasks~\cite{FewshotLL_AAAI2021,kim2023warping_ICLR2023}.

\subsubsection{Unsupervised CL}
\label{sec:unsupervisedcl}

Unsupervised CL \cite{unsuperviseCL_NeurIPS2019, madaan2022representational_ICLR2022} is a rapidly growing research area that emphasizes learning from unlabeled data alone. Unlike traditional supervised CL relying on labeled data, unsupervised CL explores techniques that enable learning and adaptation using only unlabeled data.

Existing unsupervised CL methods mainly rely on \textit{representation-based contrastive learning} techniques~\cite{unsuperviseCL_NeurIPS2019,Co2l_ICCV2021,madaan2022representational_ICLR2022,DavariUnsupervised_CVPR2022}.
CURL \cite{unsuperviseCL_NeurIPS2019} is the first offline continual
unsupervised representation learning framework with unknown task labels and boundaries. Co2l~\cite{Co2l_ICCV2021} finds that self-supervised loss is generally more robust to forgetting than cross-entropy loss in CL. 
LUMP~\cite{madaan2022representational_ICLR2022} observes that unsupervised CL models have a flatter loss landscape than supervised CL models, and additionally, it performs Mixup~\cite{mixup_ICLR2018} between old task samples and new task samples to reduce forgetting. 
Prob~\cite{DavariUnsupervised_CVPR2022} revisits the phenomenon of representational forgetting in both supervised and unsupervised CL settings, and shows that using observed accuracy to measure forgetting is a misleading metric because a low model accuracy on old tasks does not necessarily indicate significant changes in the learned representations.  This discrepancy suggests that accuracy alone is not a reliable indicator of the extent of forgetting in unsupervised CL. 

\textit{Suitable metric for unsupervised CL}: Following \cite{madaan2022representational_ICLR2022} and centered kernel alignment (CKA) \cite{kornblith2019similarity}, we measure the similarity between the representations obtained using the parameters learned at the end of task $t$ (i.e., $\vtheta_t$) and those obtained using the parameters learned at the end of the last task $T$ (i.e., $\vtheta_T$) to evaluate representation forgetting. Intuitively, a higher similarity indicates less forgetting of previous data distributions. Specifically, the similarity is calculated using test data from task $t$, as formulated below:
\begin{equation}
\centering
\small
\setlength\abovedisplayskip{3pt}
\setlength\belowdisplayskip{3pt}
    \text{sim}\left(\vtheta_t, \vtheta_T; X\right) = \frac{ \text{HSIC} (A, B)}  {\sqrt{ \text{HSIC}(A, A)  \text{HSIC} (B, B)}},
\end{equation}
where HSIC denotes Hilbert-Schmidt Independence Criterion, as defined in \cite{kornblith2019similarity}. where $(A)_{i, j} = a(\vz_i, \vz_j)$, $a(,)$ denotes a kernel function. $(A)_{i, j}$ denotes the element in $i$-th row and $j$-th column of matrix $A$. $(B)_{i, j} = b(\vu_i, \vu_j)$, where $b(,)$ denotes a kernel function.
$X$ represents the test data of task $t$. $\vz$ and $\vu$ are the representations w.r.t $X$ extracted by the CL model $f$ with weights $\vtheta_t$ and $\vtheta_T$, respectively.

CKA is more suitable for evaluating representation forgetting in unsupervised CL for the following reasons:
(1) 
CKA can capture both linear and non-linear relationships between representations.  Cosine similarity, on the other hand, is limited to linear relationships and measures only the angle between two vectors, ignoring more complex relationships.
(2) CKA is invariant to isotropic scaling and orthogonal transformations. 
Cosine similarity only accounts for the direction of the vectors and is not invariant to shifts in data, which can affect the similarity measurement if data undergoes certain transformations.

\subsection{Theoretical Analysis}

The \textit{theoretical} analysis of CL is quite a few. Pentina et al. \cite{pentina2014pac} provide a generalization bound in the PAC-Bayesian framework for CL. Karakida et al. \cite{karakida2022learning} conduct a theoretical analysis of the generalization performance within a solvable case of CL. They utilized a statistical mechanical analysis of kernel ridge-less regression to provide insights into the generalization capabilities in CL.
Kim et al. \cite{kim2022a} study class-incremental learning and provides a theoretical justification for decomposing the problem into task-id prediction and within-task prediction. 
Evron et al. \cite{CLinSeparableData_ICML2023} theoretically study the CL on a sequence of separable linear classification tasks with binary classes. Peng et al. \cite{IdeaCL_ICML2023} propose Ideal Continual Learner (ICL), which unifies multiple existing well-established CL solutions, and gives the generalization bound of ICL. Zhao et al. \cite{zhao2024statistical} statistically analyze regularization-based CL on linear regression tasks, highlighting the impact of various regularization terms on model performance.

According to \cite{IdeaCL_ICML2023}, which provides a generalization analysis for continual learner.
Each task optimizes the following learning objective:
$\gG_t := \argmin_{\vtheta \in \vTheta} \gL(\vtheta, \gD_t)$. Let $c^{*}_t = \min_{\vtheta
    } \sE_{(\vx, y) \sim \gD_t} 
    \gL(\vtheta, \vx, y)$; $d$ denotes the input dimension, i.e., $\vx \in \sR^d$; $||\vtheta||_2 \leq R$; $I_t$ denotes the number of training examples for task $t$.
\begin{assumption} \label{assume:global}
    Assume all tasks share a global minimizer, i.e., $\cap_{t=1}^{t=N} \gG_t \neq \emptyset$.
\end{assumption}
\begin{assumption} \label{assume:Lipschitz} 
    Assume the CL loss function is $K$-Lipschitz, i.e.,  $\small |\gL(\vtheta, \vx, y) - \gL(\vtheta^{\prime}, \vx, y)|\leq K ||\vtheta - \vtheta^{\prime}||_2$.
\end{assumption}
The above two assumptions ensure the following uniform convergence \cite{shalev2009stochastic} for $\delta \in (0, 1)$:
  \begin{align}
  \small
  \setlength\abovedisplayskip{3pt}
\setlength\belowdisplayskip{3pt}
        |\sum_{i=1}^{i=I_t} \gL(\vtheta, \vx_i, y_i) - \sE_{(\vx, y) \sim \gD_t} 
    \gL(\vtheta, \vx, y)| \leq \zeta(I_t, \delta)
    \end{align}
where $\small \zeta(I_t, \delta) = \gO(\frac{KR \sqrt{d\log (I_t) \log (d/\delta)}}{\sqrt{I_t}})$\;\;\; 
\begin{theorem}[\textbf{Generalization Error}]
  Assume assumption \ref{assume:global} and \ref{assume:Lipschitz} holds. With probability at least of $1-\delta$, there is the following generalization bound:
\begin{align}
\small
\setlength\abovedisplayskip{3pt}
\setlength\belowdisplayskip{3pt}
     c^{*}_t \leq \sE_{(\vx, y) \sim \gD_t} 
    \gL(\vtheta^{*}, \vx, y) \leq c^{*}_t + \zeta(I_t, \delta)
\end{align}
where $\vtheta^{*}$  denotes the optimal global CL model parameters after completing the last task.
\end{theorem}
\section{Forgetting in Foundation Model}

Forgetting in foundation models manifests in three distinct directions. (1) Firstly, when fine-tuning a foundation model, there is a tendency to forget the pre-trained knowledge, resulting in sub-optimal performance on downstream tasks. 
(2) Secondly, foundation models are typically trained on a dataset for a single pass ~\cite{PaLM,LLaMA}, resulting in two types of forgetting. In the case of streaming data, the challenge lies in retaining previous pre-trained knowledge as unlabeled data arrives sequentially \cite{jagielski2023measuring}. This type of forgetting is undesirable as it can hinder the model's ability to leverage prior knowledge effectively. Conversely, the earlier examples encountered during pre-training may be overwritten or forgotten more quickly than the later examples. While this characteristic of forgetting can be seen as a disadvantage in some contexts, it can be advantageous in privacy-preserving scenarios. Therefore, foundation models exhibit both challenges and potential benefits associated with forgetting.
(3) Lastly, foundation models have the ability to extract features with a strong feature extractor, which makes foundation models increasingly popular for CL approaches, aiming to achieve excellent performance across multiple tasks without forgetting previously learned knowledge. 
Below, we will delve into each research direction in more detail.

\subsection{Forgetting in Fine-Tuning Foundation Models}
 Fine-tuning a foundation model can result in the forgetting of pre-trained knowledge, which may lead to sub-optimal performance on downstream tasks. 
Forgetting occurs when the target model deviates significantly from the pre-trained model during the fine-tuning process ~\cite{mixout_ICLR2020}. This deviation increases the likelihood of overfitting to a small fine-tuning set ~\cite{howard2018universal_ACL2018}, which can contribute to forgetting.

There are several simple and effective strategies to mitigate forgetting during the fine-tuning process. These include techniques such as learning rate decreasing~\cite{howard2018universal_ACL2018}, weight decay~\cite{chelba2006adaptation_2006, zhang2020revisiting_ICLR2020}, and Mixout regularization~\cite{mixout_ICLR2020}. Furthermore, Fatemi et al.~\cite{GEEP_ACL2023} find that in the study of mitigating the gender bias of the pre-trained language model, the pre-trained knowledge will be forgotten when the small neutral data is fine-tuned, which will hurt the downstream task performance. Dong et al.~\cite{dong2021should_NeurIPS2021} observe that adversarial fine-tuning of pre-trained language models is prone to severe catastrophic forgetting, causing the loss of previously captured general and robust linguistic features. To address these issues, they propose a Robust Informative Fine-Tuning method from an information-theoretical perspective.
In addition, an approach called Recall and Learn, proposed in Chen et al.~\cite{chen2020recall}, addresses the forgetting issue by utilizing Pretraining Simulation and Objective Shifting. This approach enables multi-task fine-tuning without relying on the data from the pretraining tasks. Zhang et al. \cite{zhang2024dissecting} conduct a detailed analysis of forgetting in LLMs, examining its effects on the topics, styles, and factual knowledge in text.
Conjugate Prompting \cite{kotha2024understanding} seeks to reduce the forgetting of prior task knowledge during the fine-tuning of new tasks by counteracting changes in implicit task inference.
Furthermore, several recent studies have shown that merging pretrained and fine-tuned models directly at the parameter level can also help mitigate forgetting~\cite{wortsman2022robust, zhu2024model}. Moreover, recent research \cite{zhaolearning} highlights that safety fine-tuning of LLMs can lead to the forgetting of downstream task knowledge. To address this issue, ForgetFilter \cite{zhaolearning} introduces a method that filters out unsafe examples prior to fine-tuning, ensuring that downstream task performance is preserved while maintaining the safety of LLMs.

\subsection{Forgetting in One-Epoch Pre-training}

\textbf{Forgetting Previous Knowledge in Pre-training}.
Foundation model is typically trained using self-supervised learning, and self-supervised learning with \textit{streaming data} has become a significant research area due to the storage and time costs associated with storing and training on large amounts of unlabeled data. In this context, Hu et al.\cite{hu2022how} propose a sequential training method for self-supervised learning, demonstrating that self-supervised learning exhibits less forgetting compared to its supervised learning counterpart. Similarly, Purushwalkam et al.\cite{Purushwalkam0022_ECCV2022} perform self-supervised learning on a continuous non-iid data stream and introduce a minimum redundancy (MinRed) buffer approach to mitigate catastrophic forgetting. Furthermore, Lin et al.~\cite{CCL_ICME2022} develop a rehearsal-based framework that incorporates their proposed sampling strategies and self-supervised knowledge distillation to address the forgetting problem in streaming self-supervised learning.

\textbf{Beneficial Forgetting in Pre-Training}. 
The memorization and forgetting of large foundation model have attracted much attention, some emerging work \cite{jagielski2023measuring} studies the connection between the memorization of training data and different types of forgetting.
 Memorization refers to models excessively fitting to specific training examples, making them vulnerable to privacy leakage. Some researches have shown that this  phenomenon of leakage  personally identifiable information in language model \cite{lukas2023analyzing, carlini2023quantifying}, which is undesirable in practice. On the other hand, forgetting involves the gradual loss of information about examples encountered early in training. The study in \cite{jagielski2023measuring}  introduces a method to quantify the degree to which models "forget" specific details of training examples. This results in reduced susceptibility to privacy attacks on examples encountered during the early stages of model training, where these examples are typically processed for only a few epochs or even just a single epoch. Consequently, as the training progresses and numerous subsequent updates are made to the model weights, these early examples are more likely to be forgotten. This phenomenon occurs because the model's capacity is continuously adjusted to accommodate new data, causing the influence of earlier examples to diminish over time. As a result, early training examples are less vulnerable to membership inference attacks. These attacks aim to determine whether a particular example was part of the training dataset by exploiting the model's memorization to that data. However, since the early examples have a reduced impact on the model's final state, making it harder for such attacks to succeed. In essence, the transient exposure of early examples offers a form of implicit protection against these privacy threats.
Different from CL, which measures forgetting as apposed to retain knowledge from previously learned tasks. In contrast, the study conducted by \cite{jagielski2023measuring} examines a task-specific model and investigates the degree of forgetting exhibited towards specific training examples.

\subsection{CL in Foundation Model}
Recently, researchers have explored the use of foundation models to tackle continual learning (CL) problems. CL based on foundation models has shown promise. Studies such as \cite{achievingCTR_NIPS2021, wu2022pretrained} apply pre-trained models for CL in NLP tasks. Furthermore, \cite{scialom2022fine} demonstrates the capability of fine-tuned pre-trained language models to act as continual learners.

Given the remarkable success of the Transformer model ~\cite{attention_transformer_nips2017} in computer vision tasks ~\cite{vit_ICLR2021}, many studies have started exploring the application of CL using foundational model.
 Ostapenko et al. \cite{ostapenko2022continual} study the efficacy of pre-trained vision models for downstream CL tasks.
Mehta et al. \cite{Empirical_Inverstigation_Arxiv2021} explain that pretrained models help mitigate forgetting by driving weights to wider minima, which indicates better generalization.
Ramasesh et al. \cite{ramasesh2022effect_ICLR2022} observe that pretrained models are more resistant to forgetting than randomly initialized models, and this capability increases with the size of pretrained model and data.

In addition to the aforementioned studies, many researchers have explored the fine-tuning of pre-trained models to enhance their adaptation to downstream tasks in CL. These efforts involves parameter-efficient fine-tuning techniques, such as Adapters~\cite{revisitingcil_arxiv2023, CLAdapter_NeurIPS2022} and Prompts~\cite{wang2022learning_cvpr, DualPrompt_ECCV2022, sprompt_NeurIPS2022, ProgressivePrompts_ICLR2023}.
(1) {\textit{Adapter-based methods}}:
ADAM~\cite{revisitingcil_arxiv2023} demonstrates that utilizing a frozen base model to generate generalizable embeddings and setting classifier weights as prototype features can outperform state-of-the-art CL methods, with further gains from model adaptation on downstream tasks. ADA~\cite{CLAdapter_NeurIPS2022}, on the other hand, adopts a different approach by learning a single adapter for each new task instead of adjusting the entire CL model. The new adapter is then fused with the existing adapter to maintain a fixed capacity.
(2) {\textit{Prompts-based methods}}:
L2P~\cite{wang2022learning_cvpr} and DualPrompt~\cite{DualPrompt_ECCV2022} introduce a dynamic instance-level learning approach to determine the matched expert-prompt for each new task while freezing the pre-trained model. 
S-Prompt~\cite{sprompt_NeurIPS2022} focuses on exemplar-free domain-incremental learning. It independently learns prompts for each domain and stores them in a pool to prevent forgetting. Progressive Prompts~\cite{ProgressivePrompts_ICLR2023} also learns a prompt for each new task, but it freezes the prompts of old tasks and incorporates them into the new task to encourage forward knowledge transfer.

\section{Forgetting in Domain Adaptation}
\label{sec:domain_adaptation}

The objective of domain adaptation is to transfer knowledge from a source domain to a target domain. A domain represents the joint distribution of the input space $\gX$ and the output space $\gY$. Specifically, the source domain is defined as $\gP^{S}(\vx, y)$, where $\vx$ belongs to the input space $\gX^S$ and $y$ belongs to the output space $\gY^S$. Similarly, the target domain is defined as $\gP^{T}(\vx, y)$, where $\vx$ belongs to the input space $\gX^T$ and $y$ belongs to the output space $\gY^T$.

In continual domain adaptation (CDA) \cite{adaptdomain}, the focus is primarily on the covariate shift setting. Covariate shift refers to a situation where the distribution of input data, $\gX$, differs between the source and target domains, while the conditional distribution of the output, $\gY$, remains the same. This setting assumes that the relationship between inputs and outputs remains consistent across domains, but the distributions of the input data vary. This is formally defined as the following:
\begin{equation}
\small
\setlength\abovedisplayskip{3pt}
\setlength\belowdisplayskip{3pt}
    \gP^{S}(X = \vx) \neq \gP^{T}(X = \vx),  \gP^{S}(y | X = \vx) = \gP^{T}(y | X = \vx).\!\!
\end{equation}

CDA and traditional CL have distinct characteristics and goals.
On the one hand, CDA differs from traditional CL in terms of the availability of source domain data for transferring knowledge across the target domain sequence. In CDA, the source domain data is accessible, and the objective is to adapt the model from the source domain to the target domain, leveraging the available source domain data. However, the target domain may only provide unlabeled data, requiring the model to adapt to the new domain without explicit supervision.
On the other hand, traditional CL aims to learn and adapt the model to a sequence of tasks without accessing previous task-specific labeled data. Labeled data is typically provided for each task in traditional CL.

\textit{Problem Setup:}
Suppose we have a pre-trained model $f_{\vtheta}$ that has been trained on a set of source domain data $\gP^{S}(\vx, y)$, where $\vx$ belongs to the source domain input space $\gX^S$ and $y$ belongs to the source domain label space $\gY^S$. Additionally, we have a sequence of evolving target distributions $\gP_t^{T}(\vx, y)$, where $\vx$ belongs to the input space $\gX^T_t$ of the target domain $t$. $y$ belongs to the label space $\gY^T_t$ of the target domain $t$. $t$ represents the domain index ranging from 1 to $N$.
The objective of CDA, as proposed in~\cite{adaptdomain}, is to train $f_{\vtheta}$ in such a way that it performs well on all the domains $\gP_t^{T}(\vx, y)$ in the evolving target domain sequence, defined as follows: 
\begin{equation}
\small
\setlength\abovedisplayskip{3pt}
\setlength\belowdisplayskip{3pt}
    \min_{\vtheta} \mathbb{E}_{t \in [1,\ldots,N]} \mathbb{E}_{\vx^T \sim P^T_t, \vx^S \sim P^S} \mathcal{L}\left(f_{\vtheta}(\vx^T), f_{\vtheta}(\vx^S)\right).
\end{equation}
Where $\mathbb{E}$ denotes expectation,  $\mathcal{L}$ denotes the KL divergence if $f_{\vtheta}(\vx^T), f_{\vtheta}(\vx^S)$ represent the model output logits or $l_1, l_2$ distance function if they denote the feature representations.

When learning new target domains, their data distributions differ from that of the source domain. As a result, adapting the model to new domains can lead to forgetting the knowledge acquired from previous domains. Various methods have been developed to tackle the forgetting issue.

When the source domain data is available, most of the works avoid forgetting by replaying the source domain data~\cite{CUA_ICLRW2018,rostami2021lifelong,onlineDA_ECCV2022}, and a few works are based on regularization~\cite{gradient_AAAI2021}, or meta-learning~\cite{continualDA}.
First, CUA~\cite{CUA_ICLRW2018}, addresses forgetting by randomly selecting samples from previous domains and stores them in a memory buffer. UCL-GV~\cite{UCL-GV_CVPRW2022} utilizes a First-In, First-Out (FIFO) buffer to replay episodic memory.
AuCID~\cite{rostami2021lifelong} tackles CDA by consolidating the learned internal distribution. It achieves this by storing a fixed number of confident samples for each class per domain, which are later replayed during the adaptation process.
Then, GRCL~\cite{gradient_AAAI2021} utilizes the gradient direction of samples from the previous domain as a \textit{regularization} term. This constraint ensures that the model can be updated with new target domain data without negatively affecting the performance of the previous domains. 
Finally, Meta-DR~\cite{continualDA} proposes a \textit{meta-learning} and domain randomization approach to mitigate forgetting and retain knowledge from previous domains during CDA. 

Recently, few works have focused on \textit{source-free} approaches~\cite{GSFDA_ICCV2021} that protect the source domain data privacy, which is often inaccessible in many scenarios~\cite{CoSDA_Arxiv2023,C-SUDA_Arxiv2023}. CoSDA~\cite{CoSDA_Arxiv2023} introduces a knowledge distillation method that employs a dual-speed teacher-student structure. The slow-updating teacher preserves the long-term knowledge of previous domains, while the fast-updating student quickly adapts to the target domain. C-SUDA~\cite{C-SUDA_Arxiv2023}  synthesizes source-style images to prevent forgetting.

\section{Forgetting in Test Time Adaptation}

Test time adaptation (TTA) refers to the process of adapting a pre-trained model to unlabeled test data during inference or testing~\cite{wang2021tent,niu2022efficient,NOTE_NeurIPS2022,SAR_ICLR2023, liang2023comprehensive}. Unlike domain adaptation, TTA occurs during the deployment phase rather than during the training phase. 
In traditional machine learning scenarios, during testing, it is typically assumed that the test data $\gD_{test}$ follows the same distribution as the training data. However, in real-world applications, it is common for the test data distribution to deviate from the training data distribution. 
To address the distribution shift,  TTA adapts the pre-trained model on the unlabeled testing data $\vx$ using an unsupervised adaptation loss. This adaptation aims to minimize the loss function $\gL(\vx, \vtheta)$ with respect to the parameters $\vtheta$. It is important to note that $\vx$ is sampled from the testing dataset, $\gD_{test}$.
Subsequently, the adapted model utilizes the updated parameters to make predictions for the test input $\vx$. This allows the model to account for the distribution shift between training and testing, and hopefully improve its performance on  test data.

\textit{Existing Works:} 
Tent \cite{wang2021tent} minimizes the entropy of model predictions on test data, thereby improving the model's ability to generalize to unseen examples.
 MECTA \cite{hong2023mecta} proposes techniques to adapt the model during testing while optimizing memory usage, ensuring efficient and effective adaptation to the test data distribution.
MEMO \cite{zhang2022memo} applies various data augmentations to a test data point. Subsequently, all model parameters are adapted by minimizing the entropy of the model's output distribution across the augmented samples.

When a pre-trained model is adapted to new unlabeled test data, the model shifts to the new data, potentially causing it to forget crucial information previously learned from the in-distribution (ID) data. This phenomenon can result in a substantial loss of knowledge and adversely affect the model's overall performance \cite{niu2022efficient, wang2022continual}. To address this issue, existing approaches primarily adopt two strategies.

Firstly, one approach is to replay a small portion of ID data during the
adaptation process to alleviate the forgetting issue. Without loss of generality, any data selection methods mentioned in the section (memory-based methods for continual learning) can be applied. For example, one can select the sample set closest to the class center, the sample set closest to the decision boundary, or the sample set with greater diversity. RMT~\cite{dobler2023robust_CVPR2023} randomly samples 1\% of the ID data. AUTO~\cite{AUTO_Arxiv2023}  stores one sample per class, which is initialized from randomly selected training samples. During training, samples from the same class are replaced according to a predefined rule.

Secondly, one can employ a two-step process to prevent the forgetting issue. Initially, the model trained on the ID data is frozen. Subsequently, new learnable parameters are introduced to adapt the model to test data~\cite{gan2022decorate_AAAI2023,EcoTTA_CVPR2023}. For instance, VDP~\cite{gan2022decorate_AAAI2023} prevents forgetting by freezing the model trained on the ID data and instead learns a set of visual prompts tailored to the test data. These prompts help the model adapt effectively to out-of-distribution (OOD) data. Similarly, EcoTTA~\cite{EcoTTA_CVPR2023} freezes the pre-trained network on ID data and introduces a lightweight meta-network to facilitate adaptation to OOD data while retaining the valuable ID data knowledge.

Thirdly, one can constrain the updates of important parameters to prevent forgetting in TTA. Tent~\cite{wang2021tent} specifically focuses on preserving the previous knowledge by updating only the BatchNorm layer. CoTTA~\cite{wang2022continual}  randomly restores the weights of certain neurons to the weights that were originally trained on ID data. This restoration mechanism helps in retaining the knowledge acquired from the ID data.
Other approaches employ techniques similar to regularization-based approaches in traditional CL. These methods penalize the updating of parameters that are deemed important to ID data during TTA~\cite{SWR-NSP_ECCV2022,niu2022efficient,PETAL_CVPR2023}.
For instance, EATA ~\cite{niu2022efficient} calculates the importance using the Fisher information matrix to penalize the parameters updates.

\section{Forgetting in Generative Model}
\label{sec:generativemodel}
The goal of a generative model is to learn a generator that can generate samples from a target distribution. Research related to forgetting can be categorized into two categories: (1) GAN training itself can be viewed as CL;  (2) lifelong learning generative model on non-stationary distributions.

\subsection{GAN Training is a Continual Learning Problem} Thanh-Tung et al. \cite{GANforget} approach GAN training as CL. They consider the discriminator as learning a sequence of tasks, with each task representing the data distribution generated by a specific generator. To address catastrophic forgetting, they propose using momentum or gradient penalties on the discriminator. This improves convergence and reduces mode collapse.
Similarly, Liang et al. \cite{generative_Arxiv2018} adopt EWC \cite{EWC16} or SI \cite{zenke2017continual} to mitigate forgetting in the discriminator.

One application of viewing GAN training as CL is in data-free knowledge distillation (DFKD) \cite{chen2019data}. 
DFKD uses a generative model to create examples for the teacher model, but forgetting remains a significant challenge.
Forgetting occurs in DFKD due to the non-stationary distribution of pseudo-samples generated by the pseudo-data generator, as the generator evolves over time. To mitigate this forgetting issue, Binici et al. \cite{binici2022preventing, binici2022robust} use a memory buffer that dynamically collects generated samples over time or utilize generative replay to alleviate forgetting in DFKD. 
MAD \cite{MAD_NeurIPS2022}  maintains an exponential moving average of the generator, preventing drastic changes in the generated data distribution and preserving previous knowledge.
Patel et al. \cite{patel2023learning} propose a meta-learning-inspired framework to tackle the forgetting issue in DFKD.

\subsection{Lifelong Learning of Generative Models} Lifelong learning of generative models seeks to enable generative models to generate data for new tasks while preserving the ability to generate data for previously learned tasks without forgetting. The goal is to develop generative models that can continually generate high-quality samples for both new and previously encountered tasks. The proposed approaches in lifelong learning of generative models encompass both Generative Adversarial Networks (GAN)-based and Variational Autoencoders (VAE)-based methods.

In the realm of GAN-based approaches, Zhai et al. \cite{continualGAN, Zhai_2021_CVPR} have proposed lifelong GAN, which allows for conditioned image generation while avoiding the problem of forgetting previously acquired knowledge. 
In the context of VAE-based approaches, Ramapuram et al. \cite{Ramapuram_2020} introduce a lifelong learning approach to unsupervised generative modeling, continuously integrating new distributions into an existing model. This is achieved using a student-teacher Variational Autoencoder architecture, enabling the model to learn and retain all previously encountered distributions without needing to store past data or models.
Furthermore,
Ye et al. propose network expansion \cite{continualVAEmixture} and dynamic optimal transport formulation \cite{continualVAE2} for continual VAE.

\section{Forgetting in Reinforcement Learning}
\label{sec:reinforcementlearning}

While most existing CL methods primarily tackle the issue of forgetting in image classification, it is important to note that forgetting also widely occurs in reinforcement learning (RL), known as continual RL. Addressing forgetting in RL is vital for the advancement of intelligent agents that can continuously adapt to new tasks and environments \cite{khetarpal2022towards}.

Standard RL formulation could be defined as the following.
We denote $\gS$ as the state space, $\gA$ as the action space, and a reward function is $r(s_t, a_t): \gS \times \gA \rightarrow R$, where $r(s_t, a_t)$ denotes the immediate reward received after taking action $a_t$ in state $s_t$. At each time step $t$, the agent sample action from a policy function which output the optimal action or the distributions over the action space. The deterministic policy takes the current state $s_t$ as input, and outputs the optimal action $a_t$ that should be performed at the current state $s_t$ according to $a_t = \mu(s_t)$, where $\mu$ denotes the deterministic policy network. A stochastic policy takes the state $s_t$ as input, outputs the optimal action distribution according to $a_t \sim \pi(a_t|s_t)$, where $\pi$ denotes the stochastic policy network. Then, the state transition function takes the state $s_t$ and action $a_t$ as input and outputs the next state $s_{t+1}$ either deterministically $s_{t+1} = f(s_t, a_t)$ or stochastically $s_{t+1} \sim p(s_{t+1}|s_t, a_t)$, where $f$ denotes the deterministic state transition function and $p$ denotes the stochastic state transition function. The goal of RL is to accumulate as much reward as possible. Following \cite{khetarpal2022towards}, the general continual RL can be defined as:
\begin{definition} \label{continualRL}
    (General Continual RL): Given a state space $\gS$, action space $\gA$ and observation space $\gO$.  A reward function is $r: \gS \times \gA \rightarrow R$; A state transition function is $p: \gS \times \gA \rightarrow \gS $; An observation function is $x: \gS \rightarrow \gO$. The general continual RL can be formulated as
\begin{equation}
\small
\setlength\abovedisplayskip{3pt}
\setlength\belowdisplayskip{3pt}
    \gM \overset{def}{=} \langle\gS(t), \gA(t), r(t), p(t), x(t), \gO(t)\rangle.
\end{equation}
\end{definition}
 Definition \ref{continualRL} highlights that in continual RL, various components such as the state, action, reward, observation, and more, undergo changes over time. This emphasizes the dynamic nature of the RL process in continual settings.

\textbf{Continual RL approach}. The existing continual RL methods can be categorized into four main groups: (1) \textit{regularization-based methods.}  These approaches employ techniques such as knowledge distillation to alleviate forgetting \cite{igl2021transient}. Distillation can enhance experiences for training the policy or value function by offering an auxiliary target for the network to emulate. It is a widely-used technique for applying conservative updates, ensuring the agent's learning remains stable and retains essential knowledge from previous tasks. (2) \textit{rehearsal-based methods.} These methods utilize rehearsal or experience replay to mitigate forgetting \cite{daniels2022model}. Specifically, a memory buffer is employed to generate realistic samples from previous experiences. Replay techniques help reduce short-term biases in the objective function by leveraging past experiences as approximations for future situations. Therefore, replay has emerged as a highly effective approach for managing continual RL. (3) \textit{architecture-based methods.} These approaches focus on learning a shared structure, such as network modularity or composition, to facilitate continual learning \cite{gaya2023building}. CL agents need to solve problems by finding useful patterns that help them in the future. They should reuse parts of previous solutions by forming abstract concepts or skills. Humans naturally break complex tasks into smaller ones and use knowledge from different timescales to plan and learn. Equipping continual RL agents with the ability to compose relevant modules from previous experiences will help them retain and transfer knowledge. (4) \textit{meta-learning-based methods} \cite{finn17a}. Meta-learning is an effective method that boosts the learning efficiency of CL agents. By utilizing past successes and failures, the agent learns to refine its optimization processes during continual RL. If these refinements generalize effectively to future tasks, meta-learning creates an inductive bias that enhances the agent's sample efficiency and adaptability in acquiring new behaviors. This capability is crucial for achieving efficient and adaptive continual RL.

\section{Forgetting in Federated Learning}
\label{sec:federatedlearning}

Federated learning (FL) is a decentralized machine learning approach where the training process takes place on local devices or edge servers instead of a centralized server. In FL, instead of sending raw data to a central server, the model is distributed to multiple client devices or servers. Each client device performs training on its local data, and only the model updates are sent back to the central server. The central server aggregates these updates from multiple clients to update the global model. This collaborative learning process enables privacy preservation as the raw data remains on the local devices, reducing the risks associated with data sharing.

We can classify the forgetting issue in FL into two branches. The first branch pertains to the forgetting problem caused by the inherent non-IID (not identically and independently distributed) data among different clients participating in FL. In this scenario, each client's data distribution may vary significantly, leading to challenges in preserving previously learned knowledge when aggregating model updates from multiple clients.
The second branch addresses the issue of continual learning within each individual client in the FL, which results in forgetting at the overall FL level. This branch is referred to as federated continual learning (FCL) \cite{pmlr-v139-yoon21b}. FCL involves continual learning on each client while participating in FL, potentially leading to forgetting of previously learned knowledge at the global FL level. We provide a detailed explanation for each direction below.

\subsection{Forgetting Due to Non-IID Data in FL} 

The problem of \textit{client drift} \cite{karimireddy2020scaffold} in FL pertains to the scenario where there is a substantial variation in the data distribution across individual clients. In FL, each client conducts local training using its own data, and subsequently, the model updates are aggregated to construct a global model. However, when there are significant variations in the data distribution among clients, it leads to each client updating its model in different directions, ultimately resulting in a drift among the client models. This drift can lead to performance degradation or instability in the global model.
In other words, forgetting in FL occurs during the model averaging process on the server side and is primarily driven by the inconsistency of models among clients \cite{shi2023improving}. This inconsistency arises due to the presence of heterogeneous data distributions across clients.

Shoham et al. \cite{shoham2019overcoming} provide an interpretation of the client drift issue in FL by relating it to forgetting in continual learning. They propose a regularization-based approach specifically designed to address client drift in FL, particularly in the non-IID data setting. Their method aims to mitigate  forgetting and preserve previously learned knowledge during FL. 
Subsequently, multiple methods have been proposed to alleviate forgetting by integrating penalty or regularization terms that account for client model shifts ~\cite{FCCL_CVPR2022,FedReg_ICLR2022,FedNTD_NeurIPS2022}. FCCL~\cite{FCCL_CVPR2022} utilizes two teachers, the optimal model pre-trained on the client's private data and the model after the server's collaborative update, and applies knowledge distillation to constrain the client model update, thereby alleviating forgetting. FedReg~\cite{FedReg_ICLR2022} addresses forgetting by using generated pseudo data to regularize the parameter update during client training. FedNTD~\cite{FedNTD_NeurIPS2022} employs knowledge distillation to mitigate knowledge forgetting, but focuses only on distilling mispredicted classes while disregarding well-predicted ones. Additionally, inspired by the GEM~\cite{lopezpaz2017gradient} and OGD~\cite{OGDCL}, GradMA \cite{GradMA_CVPR2023} employs gradient information from previous local model and centralized model to restrict the gradient direction of the local model update. Qu et al. \cite{qu2022rethinking} propose architectural strategies to mitigate forgetting and improve the performance and stability of FL systems.

\subsection{Federated Continual Learning}

Traditional FL primarily focuses on aggregating model updates from different clients without considering the long-term retention of knowledge across multiple training rounds. However, in scenarios where clients encounter non-stationary distribution, it becomes necessary to incorporate continual learning techniques to avoid catastrophic forgetting and retain knowledge from previous training rounds.
 In standard CL, an agent learns a sequence of tasks, denoted as ${\gT_1, \cdots, \gT_N }$. However, in federated continual learning (FCL), each client learns its own private task sequence.

FCL is more complicated than traditional FL or CL because non-iid and catastrophic forgetting problems need to be solved simultaneously~\cite{pmlr-v139-yoon21b}. Solving forgetting in FCL is very challenging due to two reasons: (1) non-stationary data distribution in each client; and (2) model average in the server. Existing methods of addressing forgetting can be divided into three classes. (1) \textit{Parameter Isolation Method}~\cite{pmlr-v139-yoon21b}: FedWeIT~\cite{pmlr-v139-yoon21b} decomposes the parameters of each client into global and sparse local task adaptation parameters to reduce inter-client interference and thus alleviate forgetting.  (2) \textit{Replay-based Method}~\cite{dong2022federated,qi2023better}: inspired by traditional memory-based CL, some methods address the forgetting problem by replaying the client's old data. (3) \textit{Knowledge Distillation Method}~\cite{ma2022continual_IJCAI2022}: CFeD~\cite{ma2022continual_IJCAI2022} assumes that there is an unlabeled wild dataset in the clients and server. It proposes a distillation-based method that utilizes an unlabeled surrogate dataset to aggregate clients and avoids forgetting by rehearsing old data.

\section{Beneficial Forgetting}
\label{sec:beneficial}

Forgetting is not always harmful; in fact, intentional forgetting can prove beneficial in many cases. In this section, we explore the concept of beneficial forgetting. We begin by discussing selective forgetting and its positive impact on mitigating overfitting in Section \ref{sec:selective}. Next, we highlight the importance of discarding old, irrelevant knowledge when acquiring new knowledge in Section \ref{sec:learnnew}.
Additionally, we delve into the field of \textit{machine unlearning} in Section \ref{sec:unlearn}, which specifically focuses on the task of erasing private user data from pre-trained models to protect data privacy.

\subsection{Combat Overfitting Through Forgetting} \label {sec:selective}

Overfitting in neural networks occurs when the model excessively memorizes the training data, leading to poor generalization. To address overfitting, it is necessary to selectively forget irrelevant or noisy information~\cite{audhkhasi2019forget}. One strategy to mitigate overfitting is selective forgetting, as proposed by Shibata et al.~\cite{shibata2021learning}. This approach involves identifying and discarding less relevant or noisy information from the training data, enabling the model to focus on the most important patterns and improve generalization to unseen data. 
Current methods, including techniques such as $l_1$ normalization, feature selection, and early stopping, can be considered as forms of selective forgetting. By applying $l_1$ normalization, the model's parameters are encouraged to become sparse, effectively forgetting less relevant features. Feature selection techniques focus on identifying and retaining only the most informative features while discarding the rest. Additionally, early stopping halts the training process when the model's performance on a validation set starts to deteriorate, preventing further memorization of noise or irrelevant patterns. Furthermore, Nikishin et al.~\cite{nikishin2022primacy_ICML2022} overcome the overfitting to early experiences in reinforcement learning by resetting the last few layers of the RL agent. 

In the context of learning from noisy labeled data, the memorization and overfitting to noisy labeled data can detrimentally impact the model's generalization performance. SIGUA \cite{han2020sigua} proposes a method to selectively forget the undesired memorization of noisy labeled data and improve the generalization.

Adversarial forgetting \cite{jaiswal2020invariant} offers a mechanism to mitigate overfitting to the training data. This approach involves selectively forgetting irrelevant information to promote the learning of invariant representations, thus reducing the impact of irrelevant factors on the model's performance.

\subsection{Learning New Knowledge By Forgetting Old}  \label{sec:learnnew}

Numerous studies in psychology and neuroscience have revealed the interconnected nature of forgetting and learning \cite{gravitz2019importance_nature2019}. 
In meta-learning, the Learn to Forget approach \cite{baik2020learning} highlights that not all prior knowledge acquired through meta-learning is beneficial for learning new tasks. They propose selectively forgetting certain aspects of prior knowledge to facilitate faster learning of new tasks. Zhou et al. \cite{zhoufortuitousiclr2022} introduce a forget-and-relearn paradigm. They show that adding a forgetting step can improve the generalization and effectiveness of model relearning.
Another approach, Learning Not to Learn (LNL) \cite{kim2019learning}, focuses on forgetting biased information. LNL aims to minimize the mutual information between feature embedding and bias, thereby enhancing performance during test time by reducing the influence of biased or irrelevant information. Bevan et al. \cite{bevan2021skin} explore the idea of selectively forgetting certain biased information that might be present in the dataset to improve the accuracy and effectiveness of melanoma classification models. Chen et al. \cite{chen2022near} investigate how the selective learning and forgetting of tasks can contribute to achieving optimal task selection in meta-learning. Moreover, in continual learning, Wang et al.~\cite{wang2021afec} identify that old knowledge could interfere with the learning of new tasks.  They propose an active forgetting mechanism that selectively discards old knowledge that hinders the learning of new tasks. 

\subsection{Machine Unlearning} \label{sec:unlearn}

\subsubsection{Problem Overview}

Machine unlearning \cite{cao2015towards, machineunlearn, nguyen2022survey} addresses the need to forget previously learned training data in order to protect user data privacy and aligns with privacy regulations such as the European Union's General Data Protection Regulation \cite{mantelero2013eu} and the Right to Be Forgotten \cite{ginart2019making}. These regulations require companies and organizations to provide users with the ability to remove their data. 
Interestingly, machine unlearning demonstrates two distinct phenomena related to forgetting. On one hand, it involves forgetting specific training data memorized by the pre-trained model. The problem of machine unlearning can be categorized into two main types: exact unlearning and approximate unlearning. In Section \ref{sec:exact}, we will delve into the details of \textit{exact unlearning}, while in Section \ref{sec:approximate}, we will focus on \textit{approximate unlearning}. On the other hand, machine unlearning can lead to catastrophic forgetting of knowledge on data other than the targeted forgotten training set. The gradual updating of the pre-trained model to achieve unlearning is responsible for this second type of forgetting.  In Section \ref{sec:unlearnCF}, we will provide a comprehensive exploration of catastrophic forgetting in machine unlearning. 

\subsubsection{Exact Unlearning}  \label{sec:exact}

Exact unlearning refers to the scenario where the distribution of the model obtained by training on the remaining dataset is identical to the distribution of the unlearned model. 
Let $P(\sA(\gD))$ denote the model distribution trained on dataset $\gD$ with algorithm $\sA$. Let $\gF$ denote the dataset to be removed in the pre-trained model. $P(\sA(\gD \backslash \gF))$ denotes the model distribution by training on the remaining dataset $\gD \backslash \gF$. $P(\gU(\gD, \gF, \sA(\gD)))$ denotes the unlearned model distribution. This concept can be defined as follows.

\begin{definition}[\textbf{exact unlearning}]
For a learning algorithm $\sA$, a dataset $\gD$ and a dataset $\gF$ to be forgotten, the exact unlearning $\gU$ can be defined as following:
\begin{equation}
\small
\setlength\abovedisplayskip{3pt}
\setlength\belowdisplayskip{3pt}
    P(\sA(\gD \backslash \gF)) = P(\gU(\gD, \gF, \sA(\gD))).
\end{equation}
\end{definition}
One straightforward approach to achieve exact unlearning of targeted training data is to retrain the model using the remaining dataset. This method effectively removes all information associated with the deletion set. However, this approach can be computationally expensive, especially in the case of large-scale pre-trained models and datasets.

\textit{Existing works}. To address the above challenges, DeltaGrad \cite{pmlr-v119-wu20b} cache the model parameters and gradients at each training iteration to speed up the retraining on the remaining dataset. SISA \cite{machineunlearn}  
 involves partitioning the dataset into distinct subsets and  maintaining multiple independent models by training the model on each subset. During inference, the predictions of these individual models are  aggregated. Upon receiving deletion requests, SISA solely retrains the constituent model that originally trained on the subset containing this data point. This method ensures the elimination of all information associated with the removed data points. Sekhari et al. \cite{sekhari2021remember} explores on machine unlearning with a focus on population risk minimization instead of previous work focus on empirical risk minimization. Ullah et al. \cite{ullah2021machine} further propose an exact unlearning method based on the notion of total variation (TV) stability. ARCANE \cite{yan2022arcane} leverages ensemble learning to reduce the computational and resource requirements by transforming it into multiple one-class classification tasks.

\subsubsection{Approximate Unlearning} \label{sec:approximate}

As discussed above, exact unlearning can be computationally and memory intensive, especially in  large-scale pre-trained models and datasets. 
Additionally, it poses a challenge to validate whether the distribution of the model after unlearning matches that of a fully retrained model using the remaining dataset. As a result, recent research has embraced the concept of approximate unlearning \cite{certifiedunlearn}. This concept allows for a more efficient and feasible approach to unlearning, offering a balance between computational resources and the desired level of forgetting. The concept of "approximate unlearning" is defined as below:

\begin{definition} [\textbf{approximate unlearning} \cite{certifiedunlearn}] \label{def:approximate}
Given a $\epsilon$, for a learning algorithm $\sA$, a dataset $\gD$ and a dataset $\gF$ to be forgotten, the unlearning algorithm $\gU$ performs $\epsilon-$certified removal to remove the influence of $\vz$ defined as the following: 
\begin{equation}
\small
\setlength\abovedisplayskip{3pt}
\setlength\belowdisplayskip{3pt}
    |\log P(\sA(\gD \backslash \vz)) - \log P(\gU(\gD, \vz, \sA(\gD)))|\leq \epsilon.
\end{equation}
\end{definition}
Intuitively, the definition of approximate unlearning  strives to minimize the disparity between the parameter distribution of the unlearned model (i.e., $\gU(\gD, \vz, \sA(\gD))$) and the model obtained through full retraining on the remaining dataset (i.e., $\sA(\gD \backslash \vz)$). 
Approximate unlearning relaxes the need for an exact match, focusing on bringing distributions closer, acknowledging that exact replication on the remaining data may not always be feasible or necessary.

\textit{Existing works}. Certified Removal \cite{certifiedunlearn} introduces a  theoretical framework for approximate unlearning. It offers a compelling assurance that a model, from which specific data has been removed, is indistinguishable from a model that has never encountered that data in the first place. This theoretical guarantee provides a strong foundation for the effectiveness of the unlearning process and offers a high level of confidence in the privacy and security of the unlearning procedure.  The scrubbing procedure \cite{selectforget}  leverages a generalized and weaker form of \textit{differential privacy}. It modifies neural network weights to eliminate or "scrub" information related to specific training data. Variational Bayesian Unlearning (VBU) \cite{nguyen2020variational}  aims to minimize the KL-divergence between the approximate posterior of model parameters obtained through direct unlearning and the exact posterior from retraining the model with full data. VBU strikes a balance between completely forgetting the erased data and retaining essential knowledge captured by the model's posterior belief when trained on the full data. L-CODEC \cite{mehta2022deep} identifies a specific subset of model parameters with the highest semantic overlap on an individual sample level. This subset of parameters is then utilized to achieve more efficient and targeted unlearning.

\textit{Connection with Differential Privacy}: 
Approximate unlearning is closely related to differential privacy \cite{dwork2006calibrating}, which is defined as the following:

\begin{definition}[\textbf{differential privacy} \cite{dwork2006calibrating}] 
A (randomized) algorithm 
$\sA$ is said to be 
$\epsilon$-differential private if for any two dataset $\gD$ and $\gD^{\prime}$ that differ in exactly one element:
\begin{equation}
\small
\setlength\abovedisplayskip{3pt}
\setlength\belowdisplayskip{3pt}
    |\log P(\sA(\gD)) - \log P(\sA(\gD^{\prime}))|\leq \epsilon.
\end{equation}
\end{definition}
where lower values of $\epsilon$ indicates stronger privacy. If we set $\gD^{\prime} = \gD \backslash \vz$, it is evident that differential privacy (DP) of the algorithm $\sA$ is a sufficient condition for the $\epsilon$-approximate unlearning simply by setting the unlearning algorithm $\gU$ to be identity function \cite{certifiedunlearn} in Definition \ref{def:approximate}. On the other hand, DP is not a necessary condition to achieve unlearning because there are other methods to achieve the same goal without the privacy guarantees that DP provides. Unlearning can be achieved through retraining on remaining datasets, direct unlearning algorithms, or other specialized techniques that do not rely on the noise addition and bounded influence principles of DP. Therefore, while DP provides stronger privacy guarantees, it often requires adding noise to the model or its outputs, which can lead to a decrease in model accuracy and utility. In contrast, approximate unlearning may not offer the same rigorous privacy guarantees as DP but aims to balance forgetting specific data with retaining overall performance. Approximate unlearning may provide a more practical trade-off between privacy and utility.

\subsubsection{Catastrophic Forgetting Other Normal Examples} \label{sec:unlearnCF}

Catastrophic forgetting in machine unlearning refers to the unintended decrease in the log posterior probability of the remaining dataset, denoted as $\small P(\vtheta \mid \gD\backslash\gF)$, as a result of unlearning \cite{anomalydetection}. This phenomenon leads to a degradation in the performance of the unlearned model on data other than the targeted samples, which is undesirable. The significance of catastrophic forgetting increases as more data is unlearned. It is important to note that this concept differs from the objective of machine unlearning, which focuses solely on forgetting specific samples. UNLEARN \cite{anomalydetection} constrains the update magnitude of new parameter values to the old values, and a memory buffer is maintained to store previous examples. The Forsaken method \cite{ma2022learn} introduces a dynamic gradient penalty term to restrict parameter changes on normal data, effectively mitigating the issue of catastrophic forgetting. VBU \cite{nguyen2020variational} provides a natural trade-off interpretation between fully unlearning from erased data and retaining posterior beliefs by learning on the entire dataset. Additionally, scrubbing procedure \cite{selectforget} minimizes loss on the remaining data while maximizing forgetting on the target forgetting dataset by introducing random noise to the model weights, allowing for forgetting on the specified dataset while retaining knowledge on the remaining data. Furthermore, PUMA \cite{wu2022puma} reweighs the remaining data, ensuring that the model's performance is not excessively affected by unlearning. 
Finally, it is evident that we cannot simply erase an excessive number of data points, as doing so would eventually lead to a decrease in performance on the test set.  Sekhari et al. \cite{sekhari2021remember} analyze how many samples can be deleted from a pre-trained model without affecting test performance, offering insights into the trade-off between unlearning and generalization.

\subsubsection{Application of Machine Unlearning}

Machine unlearning techniques find application in various domains, including backdoor attack defense, defending against membership inference attacks and continual learning. In the backdoor attack defense \cite{liu2022backdoor}, machine unlearning can be used to erase the backdoor trigger that has been injected into a trained model. By selectively unlearning the specific information related to the backdoor trigger, the model can be cleansed of its malicious behavior.
Similarly, machine unlearning can also serve as a defense mechanism \cite{wu2022puma, certifiedunlearn} against membership inference attacks. By removing the specific training data from a pre-trained model, the traces of these data are eliminated, making it harder for attackers to infer membership or extract sensitive information. In continual learning, machine unlearning can enhance the retention of previously learned knowledge \cite{wang2024a}.

\textit{Intentional Forgetting in Diffusion Model:}  
The advancement of text-to-image diffusion models \cite{ho2020denoising, saharia2022photorealistic} has sparked significant concerns regarding data privacy, copyright infringement, and safety related to generative models, primarily because of the memorization effect observed in large generative models. To address these concerns, the Forget-Me-Not approach \cite{zhang2023forget} and Selective Amnesia (SA) \cite{heng2023selective} aim to mitigate the presence of potentially harmful or unauthorized content. These approaches achieve this by leveraging forgetting mechanisms/machine unlearning to eliminate unwanted information.

\section{Discussion and Future Prospect} \label{sec:discussion}

\subsection{Summary and Research Trends}
The current research trends are summarized as follows:
(1) \textit{Cross Discipline Research About Forgetting: }
This exploration encompasses both the application of CL techniques to address challenges in other research domains and applications of techniques from various fields to solve problems in CL. 
In future work, we anticipate that actively promoting collaboration and facilitating idea exchange across diverse research areas will be crucial in overcoming disciplinary barriers.
(2) \textit{Intentional Forgetting: }
Intentional forgetting has emerged as a promising approach to enhance model performance and address data privacy concerns in recent studies. 
(3) \textit{Growing Trend for Theoretical Analysis: } 
Recently, there is an emerging trend towards conducting theoretical analyses to delve deeper into the understanding and analysis of forgetting across diverse research fields. These theoretical analyses aim to provide more valuable insights and uncover fundamental principles that govern forgetting.

\subsection{Open Research Questions and Future Prospect}

We prospect several potential future research directions: 
(1) \textit{In-depth Theoretical Analysis of Forgetting: }
The majority of existing methods are empirical in nature, lacking theoretical guarantees and comprehensive analysis. Therefore, there is a pressing need for more thorough theoretical analysis across different learning domains. 
(2) \textit{Proper Trade-off Between Memorization and Forgetting: } \label{sec:tradeoffmemoryforgetting}
Remembering more previous knowledge can increase the risk of memorizing private information, potentially compromising privacy. Conversely, prioritizing privacy protection may result in sacrificing performance on previous tasks. Future research should focus on developing methodologies and techniques that enable effective management of knowledge retention while ensuring privacy preservation, thereby striking the optimal balance between the two.

\section{Conclusion}

Forgetting is a prevalent phenomenon across various machine learning fields, driven by different factors. This survey aims to offer a comprehensive examination of the forgetting issue in diverse machine learning domains.
Our survey aims to provide a thorough overview and understanding of the research progress on forgetting. While existing surveys on continual learning often emphasize the harmful aspects of forgetting, our survey argues that forgetting can also be beneficial.
We present application scenarios where selective forgetting improves generalization performance and learns new tasks, and intentional forgetting of memorized examples helps safeguard the privacy of machine learning.

\IEEEdisplaynontitleabstractindextext

\IEEEpeerreviewmaketitle

% \appendices

% \section{}
% Appendix two text goes here.

% use section* for acknowledgment
%\ifCLASSOPTIONcompsoc
  % The Computer Society usually uses the plural form
%  \section*{Acknowledgments}
%\else
%  % regular IEEE prefers the singular form
%  \section*{Acknowledgment}
%\fi

{\small
\bibliographystyle{IEEEtran}
\bibliography{Reference}
}

\ifCLASSOPTIONcaptionsoff
  \newpage
\fi

\clearpage
\appendices

% you can choose not to have a title for an appendix
% if you want by leaving the argument blank
\section{Forgetting in Continual Learning}

\subsection{Task-aware CL} 
\label{sec:taskawarecl_appendix}

Existing works proposed for solving task-aware CL have five major research branches: memory-based, architecture-based, regularization-based, subspace-based, and Bayesian-based CL methods. We give an overall framework in Fig.~\ref{fig:cl_classification}. We present the details of each class method in the following.

\paragraph{\textbf{Memory-based Method}} 

Memory-based method keeps a \emph{memory buffer} that stores the data examples from previous tasks and replay those examples during learning new tasks, representative works include \cite{generativereplay,AGEM19,riemer2018learning, ERRing19, NIPS2019_9357}. It can be further categorized into: (1) raw memory replay; (2) memory sample selection; (3) generative replay; (4) compressed memory replay. Next, we discuss each direction in detail.

\textit{Raw Sample Replay:} These methods save a small amount of data from previous tasks and trains the model with the new task.
A key issue in this type of approach is how to select raw samples to store in the memory buffer. Current research works contain random selection~\cite{ERRing19} and selection based on heuristic rules~\cite{chaudhry2018riemannian,NIPS2019_9357}.
The \textit{random selection} method randomly saves some raw samples into the memory, and performs replay~\cite{ERRing19} or constraints~\cite{lopezpaz2017gradient,AGEM19} when learning a new task. For example, 
GEM~\cite{lopezpaz2017gradient} and A-GEM~\cite{AGEM19} use the memory buffer data losses as inequality constraints to
constrain their increase but allow their decrease to mitigate forgetting. Meta Experience Replay (MER)~\cite{riemer2018learning} uses meta-learning to encourage information transfer from previous tasks and minimize interference. DER++~\cite{buzzega2020dark} found that replaying old data with soft labels is significantly better than one-hot labels.

\textit{Memory Sample Selection:} These simplest methods to mitigate forgetting is to randomly store some examples for replay. However, this may lead to sub-optimal performance since random selection overlooks the informativeness of each sample. The \textit{heuristic selection} selects samples for storage based on certain rules. e.g., iCaRL~\cite{icarl} selects the samples closest to each category cluster center. 
Some works choose the samples closest to the decision boundary of each class~\cite{chaudhry2018riemannian,NIPS2019_9357} or make the sample diversity in memory the highest~\cite{aljundi2019gradient,bang2021rainbow}. GCR~\cite{GCR_CVPR2022} selects the corset that best represents the gradients produced by past data.
Sun et al.~\cite{InfluenceCL_CVPR2023} model and regularize the influence of each selected example on the future.
In addition, the above schemes of static memory allocation (fixed total capacity or storing a specific number of samples per class) may also lead to suboptimal performance. RMM~\cite{RMM_NeurIPS2021} proposes a reinforcement learning based strategy to automatically allocate capacity for old and new classes.

\textit{Generative Replay:} When privacy concerns restrict the storage of raw memory data, generative replay provides an alternative approach in CL to replay previous task data. The main concept behind generative replay is to train a generative model capable of capturing and remembering the data distribution from previous tasks. Instead of storing raw samples, the generative model is used to generate synthetic samples representative of the old tasks ~\cite{LAMOL_ICLR2020}. The representative works in this line involve using different generative models, include GAN-based methods~\cite{chenshen2018memory,xiang2019incremental}, AutoEncoder-based methods~\cite{kemkerfearnet}, Diffusion-based model~\cite{cil_diffusion_Arxiv2023}, and Model-inversion~\cite{AlwaysBeDreaming_ICCV2021}.

\textit{Compressed Memory Replay:} In scenarios where storage constraints are stringent on edge devices, memory efficiency becomes a crucial consideration. Several research efforts have aimed to improve memory efficiency in CL by employing different strategies. For example, instead of directly storing high-resolution images, one approach is to store \textit{feature representations} ~\cite{van2020brain_naturecommunications2020,iscen2020memory,BiRT_ICML2023}  or \textit{feature prototype} extracted from the image ~\cite{zhu2021prototype,SSRE_cvpr2022}, or compress the high-fidelity image into a low-fidelity image (JPEG) for storage~\cite{wang2022memory,Luo2023ClassIncremental}. 
Recently, dataset condensation has emerged as a promising approach for compressing and condensing datasets, as demonstrated by works such as \cite{zhao2021dataset, cazenavette2022dataset}. Drawing inspiration from data distillation techniques, many CL methods adopt the strategy of storing \textit{condensed samples} from old tasks in memory for replay ~\cite{Mnemonics_liu2020mnemonics,dengremember2022}.

\paragraph{\textbf{Architecture-based Method}}  Architecture-based methods in CL \cite{rusu2016progressive_pnn, fern2017pathnet, yoon2018lifelong} involve updating the network architecture during the learning process to retain previously acquired knowledge. These methods aim to adapt the model's architecture to accommodate new tasks while preserving the knowledge from previous tasks.
Based on whether the model parameters expand with the number of tasks, architecture-based methods can be categorized into two types: fixed-capacity and capacity-increasing methods.

\textit{Fixed-Capacity}:
In these methods, the amount of CL model's parameters does not increase with the number of tasks, and each task selects a sub-network from the CL model to achieve knowledge transfer and reduce the forgetting caused by sub-network updates. Common subnetwork selection techniques include masking~\cite{HAT2018,bellecdeep,wangsparcl2022}, and pruning~\cite{mallya2018piggyback,mallya2018packnet,hung2019compactingNeurIPS}. 
On the one hand, the \textit{mask-based} method isolates the update of important neurons or parameters of old tasks during the backpropagation of the new task, thereby alleviating the forgetting of old tasks.
A representative method, HAT~\cite{HAT2018} proposes a task-based binary/hard attention mask mechanism on neurons instead of parameters. Unlike HAT, which requires task identification during testing, SupSup~\cite{wortsman2020supermasks} proposes a gradient-based optimization strategy, which can automatically infer tasks and find the learned optimal mask during testing. In addition, different from the above-mentioned hard mask methods that monopolize the sub-network, which hinders knowledge transfer and has the problem of excessive consumption of network capacity by old tasks, SPG~\cite{SPG_ICML2023} proposes a soft mask mechanism that uses the importance of old task parameters as a mask to constrain the back-propagation gradient flow updates to important parameters. 
On the other hand, similar to the mask mechanism, the \textit{pruning}-based method freezes important parameters to avoid forgetting after the old tasks are learned, and reinitializes unimportant parameters to learn new tasks.
PackNet~\cite{mallya2018packnet} iteration performs pruning of a well-trained base network and maintains the binary sparse mask. Similarly, WSN~\cite{kang2022forget} selects an optimal sub-network (winning ticket) for each task based on the Lottery Ticket Hypothesis~\cite{franklelottery2019}. 
Finally, unlike all the above methods that start masking or pruning from a densely connected network, although NISPA~\cite{gurbuz2022nispa} also fixes the total number of neurons, it only starts learning tasks from a sparsely connected network, and then continuously increases the connections between neurons to learn new tasks.

\textit{Capacity-Increasing}: 
a fixed capacity CL model may face limitations in accommodating new tasks as the number of tasks increases. This can restrict the plasticity of new tasks due to insufficient remaining network capacity. To overcome this challenge, dynamic capacity methods have been proposed. These methods augment the CL model by introducing task-specific new parameters for each new task while freezing the parameters associated with old tasks. By doing so, dynamic capacity methods aim to prevent forgetting while allowing the model to adapt and learn new tasks effectively~\cite{rusu2016progressive_pnn,yoon2018lifelong,L2grow,Der_CVPR2021}. For example, PNN~\cite{rusu2016progressive_pnn} employs an independent network branch for each task, where the input of the new task branch is the output from the old task branch, enabling knowledge transfer between tasks. 
ExpertGate~\cite{ExpertGate_CVPR2017}, MEMO~\cite{zhou2023model}, and DNE~\cite{DNE_CVPR2023} also add task-specialized experts/blocks/layers for each new task, respectively. 
BatchEnsemble~\cite{batchensemble_ICLR2020} freezes the backbone network after learning the first task, and learns two independent rank-one matrices for each new task. However, it also limits knowledge transfer between tasks starting from the second task.

 \paragraph{\textbf{Regularization-based Method}} 
Regularization-based methods~\cite{zhao2024statistical} in CL involve the addition of regularization loss terms to the training objective to prevent forgetting of previously learned knowledge \cite{EWC16,zenke2017continual,oswald2019continual}. These methods aim to constrain the model's parameter updates during the learning process to ensure that valuable information from past tasks is retained.
Regularization-based methods can be further divided into two subcategories: penalizing important parameter updates and knowledge distillation using a previous model as a teacher.

\textit{Penalize Parameter Updates:} Some works alleviate forgetting by penalizing parameter updates that are important for old tasks. For example, EWC~\cite{EWC16} calculates regularization terms by approximating Fisher information matrix (FIM). MAS~\cite{MAS2018} takes the cumulative update of the parameters as a penalty term. SI~\cite{zenke2017continual} computes parameter importance using the path integral of gradient vector fields during parameter updating. Rwalk~\cite{chaudhry2018riemannian} can be seen as a generalized version of SI and EWC++.
UCL~\cite{UCL_NeurIPS2019} proposes an uncertain regularization based on the Bayesian online learning framework. AGS-CL~\cite{AGS-CL_NeurIPS2020} proposes two group sparsity penalties based on node importance as regularization terms.

\textit{Knowledge-Distillation-Based: } 
Inspired by knowledge distillation \cite{hinton2015distilling}, several methods in CL incorporate a distillation loss between the network of the previous task (referred to as the teacher) and the network of the current task (referred to as the student) to mitigate forgetting.
One method that follows this approach is Learning without Forgetting (LwF) \cite{LwF}, which considers the soft targets generated by the teacher network as additional learning objectives for the student model. By treating the soft targets as valuable information, LwF enables the current model to learn from the knowledge of the previous model, reducing the risk of forgetting.
Building upon LwF, LwM \cite{dhar2019learninglwm} leverages the attention mechanism of the previous network to guide the training of the current network. The attention of the teacher network helps the student model focus on relevant features and retain important knowledge from past tasks.
Another method, BMC \cite{bmc_cvpr2023}, takes a different approach by training multiple expert models simultaneously on disjoint sets of tasks. These expert models are then aggregated through knowledge distillation, which involves transferring knowledge from the expert models to a single student model. 
On the other hand, in distillation-based CL methods, the ideal scenario would involve using the raw data of old tasks to extract the knowledge of the teacher model and distill it to the student model. However, accessing the raw data of old tasks is often not feasible due to data privacy concerns. Consequently, existing approaches utilize proxy data as a substitute for distillation.
One such method is LwF \cite{LwF}, which uses the current task data as a proxy for distillation. 
Another approach, GD \cite{lee2019overcoming}, utilizes large-scale unlabeled data from the wild as a proxy for distillation.

\paragraph{\textbf{Subspace-based Method}} 
\label{sec:subspace_appendix}
Subspace-based methods in CL aim to address the issue of interference between multiple tasks by conducting learning in separate and disjoint subspaces~\cite{qiao2024prompt,xiao2024hebbian}. Subspace-based methods can be categorized into two types based on how the subspaces are constructed: orthogonal gradient subspace and orthogonal feature subspace methods.

\textit{Orthogonal Gradient Subspace}:
These methods require that the parameter update direction of the new task is orthogonal to the gradient subspace of the old tasks, ensuring minimal interference between tasks. OGD~\cite{OGDCL} proposes updating the CL model in a gradient subspace orthogonal to the subspace of the old tasks.
PCAOGD~\cite{PCAOGD_AISTATS2021} is an extension of OGD that only stores the top principal components of gradients for each task. ORTHOG-SUBSPACE~\cite{ChaudhryOrthogSubspaceCL} learns different tasks in different orthogonal (low-rank) subspaces to minimize interference.

\textit{Orthogonal Feature Subspace}:
These require that the parameter update direction of the new task is orthogonal to the subspace crossed by the input (feature) of the old task.
OWM~\cite{owm} proposes that when learning a new task, network parameters are only updated in a direction orthogonal to the subspace spanned by the inputs of all previous tasks. AOP~\cite{guo2022adaptive} solves the problem of inaccurate input space estimation in OWM.
GPM~\cite{saha2021gradient} proposes to construct subspace for past tasks, the network updates the network parameters taking gradient steps
in the orthogonal direction to the gradient subspaces that are important for the
past learned tasks. 
Deng et al.\cite{deng2021flattening} propose a Flattening Sharpness for Dynamic Gradient Projection Memory (FS-DGPM) with soft weight on the basis of GPM to improve new task learning. 
TRGP \cite{lin2022trgp} introduces a trust-region variant of GPM to tackle the challenge of balancing the learning of new tasks with preserving the knowledge of old tasks.

\textbf{Discussion}.
In the following, we discuss the respective advantages and disadvantages of gradient projection in feature space and gradient space, as well as how to choose them.
%
% \noindent
(i) \textbf{\textit{Gradient Projection in Feature Space.}}
\textit{Advantages}:  Projecting gradients into the feature space, which is often lower-dimensional than the gradient space, reduces computational and memory complexity.
\textit{Disadvantages}: (1) The effectiveness is highly dependent on the quality and relevance of the feature space. Poor features can lead to suboptimal performance. (2) The fixed low dimensional feature space may limit the model's ability to adapt to new tasks that require different features.
%
% \noindent
(ii) \textbf{\textit{Gradient Projection in Gradient Space.}}
\textit{Advantages}: (1) Directly operating on gradients allows for precise adjustments specific to each task. (2) It is more flexible to adapt to diverse tasks, as it directly manipulates the parameter updates without being constrained by a fixed feature space.
\textit{Disadvantages}: Gradient space can be high-dimensional, leading to higher computational and memory costs.
%
% \noindent
(iii) \textbf{\textit{When to Choose Which?}}
Feature space projection is suitable when computational and memory efficiency  are crucial. On the other hand, gradient space projection is suitable for applications requiring high flexibility and precise adaptation to new tasks.

\paragraph{\textbf{Bayesian Method}} 
\label{sec:appendix_bayesian}
Bayesian methods provide a principled probabilistic framework for addressing CF and can be classified into three categories: (1) methods that constrain the update of weight parameter distributions, (2) methods that constrain the update in function space, and (3) methods that dynamically grow the CL model architecture in an adaptive and Bayesian manner. These Bayesian approaches offer effective strategies to mitigate CF by incorporating uncertainty estimation and regularization techniques, thereby enhancing the adaptability of the learning process. In the following paragraphs, we provide a detailed discussion of each research direction.

\textit{Weight Space Regularization:} Weight space regularization based methods model the parameter update uncertainty and enforce the model parameter (weight space) distribution when learning the new task is close to that of all the previously learned tasks, including \cite{nguyen2017variational,IMM_NeurIPS2017,Adel2020Continual}. Gaussian
Residual Scoring (GRS) \cite{Kurle2020Continual} extends Bayesian neural network to the non-stationary streaming data. Different from previous work, which updates the parameter posterior distribution recursively over the task sequence, posterior meta-replay \cite{henning2021posterior} which learns 
the task-specific posteriors via a single shared meta-model, improving the flexibility in representing the current task and previous tasks. Natural Continual Learning (NCL) \cite{kao2021natural} combines Bayesian inference with gradient projection methods.

\textit{Function Space Regularization:} Different from weight space regularization which constrains the weight update, the function space regularization regulates the CL function update in the function space. Functional Regularisation of Memorable Past (FROMP) \cite{pan2020continual} enforce the posterior distribution over function space (instead of weight space) of the new task to be close to that of the all the previously learned task with Gaussian process. In contrast to previous works that primarily emphasize either parameter or function space regularization, Functional Regularized Continual Learning (FRCL) \cite{titsiasfunctional2020} takes a different approach by focusing on constraining neural network predictions to prevent significant deviations from solutions to previous tasks. Variational Auto-Regressive Gaussian
Processes (VAR-GPs) \cite{kapoor2021variational} further proposes a framework through  a principled posterior updating mechanism to naturally model the cross-task covariances.
Sequential function-space variational inference (S-FSVI) \cite{rudner2022continual} presents a framework for addressing CL by formulating it as sequential function-space variational inference. This approach offers several advantages, including the ability to employ more flexible variational distributions and more effective regularization techniques.

\textit{Bayesian Architecture Expansion:} Bayesian architecture growing methods employ a probabilistic and Bayesian approach to dynamically expand the CL model. By leveraging Bayesian principles, these methods enable the CL model to incrementally grow and adapt to new tasks or data while preserving previously learned knowledge. This probabilistic framework facilitates the flexible and principled expansion of the model's architecture, allowing it to accommodate increasing complexity and variability in the learning process. Kumar et al. \cite{kumar2021bayesian} introduce a principled framework for CL that incorporates both variational Bayes and a nonparametric Bayesian modeling paradigm. Their approach enables the continual learning of neural network structures, offering a systematic methodology for adapting and expanding the network architecture over time. Moreover, the work by Mehta et al. \cite{mehta2021continual} introduces IBP-WF, a method that leverages a principled Bayesian nonparametric approach, to dynamically expand the network architecture based on the task complexity. This approach allows the model to scale and accommodate increasing demands and intricacies of the learning tasks, ensuring efficient and effective adaptation throughout the CL process. 

\textbf{Discussion}.
{Difference and connection between weight-space and function-space regularization}:
(1) \textit{Weight-space regularization} focuses on penalizing changes in \textit{model parameters}, interpreted in Bayesian terms as constraining the posterior changes over weights.
(2) \textit{Function-space regularization} focuses on penalizing changes in \textit{model outputs}, aligned with maintaining a posterior over functions in a Bayesian framework. 
(3) \textit{Comparisons}: Simply keeping current weights close to previous ones doesn't always guarantee that predictions on past tasks will stay the same. Given the complex relationship between weights and network outputs, ensuring the effectiveness of weight regularization can be challenging.
Regularizing model outputs in function space is more effective for maintaining consistent outputs but may be more computationally costly than doing so in weight space.

\subsection{Few-shot CL}
\label{sec:fewshotcl_appendix}

In order to tackle the forgetting problem within the context of few-shot CL, existing approaches employ various techniques, including metric learning, meta-learning, and parameter regularization. These techniques will be elaborated upon in the following.

\paragraph{\textit{Metric Learning-Based}} These methods perform classification by class prototypes. To avoid forgetting, the prototype of the new class should be separable from the old class~\cite{zhu2021self_cvpr}, and the prototype of the old class should not change drastically during the adjustment process of the new class~\cite{yangneural_2023_iclr}.
TOPIC~\cite{tao2020few_cvpr2020} first defines the benchmark-setting of few-shot class-incremental learning (FSCIL), which represents the topological structure of different classes in the feature space, and alleviates the forgetting of old classes by keeping the topology stable. To alleviate forgetting, DSN~\cite{yang2022dynamic_tpami} estimates the distribution of classes in the current task. In a new task, feature vectors are sampled from old class distributions for replay. 
Another two approaches, C-FSCIL ~\cite{hersche2022constrained_cvpr} and FSCIL~\cite{yangneural_2023_iclr} try to reduce interference between prototypes. C-FSCIL maps input images to quasi-orthogonal prototypes to minimize task interference. Also, FSCIL pre-assigns and fixes the feature prototype classifier, and trains a projection layer to project the sample features of each class onto the assigned prototype, thereby avoiding the interference between classes and the problem of forgetting.

\paragraph{\textit{Meta-Learning-Based}} These methods simulate the inference phase during training so that CL models can quickly adapt to unseen new classes to solve few-shot CL. LIMIT~\cite{zhou2022few_tpami} and MetaFSCIL~\cite{chi2022metafscil_cvpr} split the base task into multiple 'fake'-incremental tasks, so that the model has the learning ability of FSCIL tasks. Specifically, MetaFSCIL directly considers the adaptability of new tasks and the stability of old tasks as its primary objectives for meta-learning. When MetaFSCIL trains on a new class, it evaluates the performance on all encountered classes as a meta-objective. By reducing the loss associated with the meta-objective, MetaFSCIL minimizes forgetting of the old tasks.

\paragraph{\textit{Parameter Regularization-Based}} These methods employ various strategies to address the forgetting problem by penalizing parameter updates that are important for old tasks. For example, FSLL~\cite{FewshotLL_AAAI2021} and WaRP~\cite{kim2023warping_ICLR2023} adopt an approach where certain crucial parameters are frozen during training, while the remaining parameters are fine-tuned specifically for the few-shot task. 
Another approach, Subspace Regularization~\cite{akyureksubspace_2022_iclr}, introduces a subspace regularization that encourages the weights of new classes to be close to the subspace spanned by the weights of existing classes.

\section{Forgetting in Meta-Learning}
\label{sec:metalearning}

Meta-learning, also known as learning to learn, focuses on developing algorithms and models that can learn from previous learning experiences to improve their ability to learn new tasks or adapt to new domains more efficiently and effectively.
In meta-learning, the goal is to enable a learning system, often referred to as the meta-learner or the meta-model, to acquire general knowledge or "meta-knowledge" from a set of related learning tasks or domains. This meta-knowledge is then leveraged to facilitate faster learning, better generalization, and improved adaptation to new, unseen tasks or domains.

Formally, let's consider a distribution of tasks, denoted as $P(\gT)$. For a specific task $\gT_t$, which consists of a training dataset $\gD_{\text{train}}$ and a validation dataset $\gD_{\text{val}}$, sampled from the task distribution $P(\gT)$.
The loss function for task $\gT_t$ with meta-parameters $\vtheta$ is defined as:
\begin{equation}
\small
\gL(\gT_t) = \log P(\gD_{\text{val}}|\gD_{\text{train}};\vtheta).
\end{equation}
The objective of meta-learning is to optimize the meta loss function, given by:
\begin{equation}
\small
\min_{\vtheta} \mathbb{E}_{\gT_t \sim P(\gT)} \gL(\gT_t).
\end{equation}
In other words, the aim is to find the optimal meta-parameters $\vtheta$ that minimize the expected loss across tasks, where tasks are sampled from the task distribution $P(\gT)$.

However, forgetting can still occur in the context of meta-learning, and it can be classified into two distinct research directions.
The first research direction focuses on Incremental Few-Shot Learning (IFSL), where the objective is to meta-learn new classes in addition to the pre-trained base classes. In this scenario, forgetting arises from the loss of information related to the pre-trained base classes. The challenge lies in retaining the knowledge of both the base classes and the newly introduced classes during the learning process.
The second research direction deals with Continual Meta-Learning, where the agent encounters non-stationary task distributions over time while learning new tasks. Unlike IFSL, the goal here is not to remember the specific base classes. Instead, the objective is to retain the meta-knowledge acquired from previous task distributions. We will present the details of each direction in the following.

\subsection{Incremental Few-Shot Learning}

Incremental few-shot learning (IFSL) \cite{gidaris2018dynamic, ren2018incremental} focuses on the challenge of learning new categories with limited labeled data while retaining knowledge about previously learned categories. 
In this scenario, a standard classification network has previously undergone training to recognize a predefined set of base classes. After that, the focus is on incorporating additional novel classes, each accompanied by only a small number of labeled examples. Subsequently, the model is tested on its classification performance, considering both the base and novel classes.

\textit{Existing Works:} Gidaris et al.  \cite{gidaris2018dynamic} propose the IFSL problem and an attention-based solution to mitigate the forgetting in IFSL.   
The Attention Attractor Network, proposed by Ren et al. \cite{ren2018incremental}, is an alternative approach where the per-episode training objective during the incremental meta-learning stage is regulated using an attention mechanism to attend the set of base classes. In contrast to previous approaches that extract a fixed representation for each task, XtarNet \cite{yoon2020xtarnet} emphasizes the extraction of task-adaptive representations by combining novel and base features to enhance the adaptability of the representations.
Shi et al. \cite{shi2021overcoming} suggest putting more effort into the base classifier pretraining stage rather than the later few-shot learning stage. As a result, they propose to seek flat local minima of the base classifier training objective function and subsequently fine-tune the model parameters within that flat region when faced with new tasks. 
In addition, C-FSCIL \cite{hersche2022constrained_cvpr} incorporates a trainable fixed-size fully connected layer and a rewritable dynamically growing memory buffer to mitigate forgetting. This memory buffer can store a vector for each class encountered up to that point in the learning process.

\subsection{Continual Meta-Learning}

The goal of continual meta-learning (CML) is to address the challenge of forgetting in non-stationary task distributions. Traditional meta-learning approaches typically focus on a single task distribution. However, CML extends this concept to handle a sequence of task distributions, denoted as $P_1(\gT), P_2(\gT), \cdots, P_N(\gT)$. 
In CML, the objective is to develop meta-learning algorithms that can effectively adapt and generalize to new task distributions as they arise over time. These task distributions can represent different environments, domains, or contexts. It aims to mitigate the forgetting of previously learned task distributions while efficiently adapting to new tasks.

\textit{Existing Works:} 
Online meta-learning (OML) \cite{finn2019online} is a framework that assumes tasks arrive sequentially and aims to improve performance on future tasks. Jerfel et al. \cite{jerfel2018reconciling} extended the Model-Agnostic Meta-Learning \cite{finn17a} approach and utilized Dirichlet process mixtures to group similar training tasks together. However, this method is not scalable to large-scale non-stationary distributions due to the requirement of independent parameters for each component.
Yap et al. \cite{continualFS} proposed an approach to model the posterior distribution of meta-parameters using Laplace approximation \cite{nguyen2017variational}.  Zhang et al. \cite{zhang2021variational} further extended this framework by employing a dynamical mixture model to learn the distribution of meta-parameters instead of a single distribution. Additionally, they used structural variational inference techniques to infer latent variables in the model.
Wang et al. \cite{Wang_2022_CVPR, Wang_2021_ICCV, Wang_2022_ECCV} introduced a large-scale benchmark for sequential domain meta-learning. They proposed different settings, including supervised learning, imbalanced domains, and semi-supervised settings, to evaluate the performance of various methods in sequential domain meta-learning.

\end{document}